\documentclass[10pt,twocolumn,letterpaper]{article}

\usepackage{cvpr}              
\definecolor{cvprblue}{rgb}{0.21,0.49,0.74}
\usepackage[pagebackref,breaklinks,colorlinks,allcolors=cvprblue]{hyperref}

\usepackage{graphicx}
\usepackage{amsmath}
\usepackage{amssymb}
\usepackage{booktabs}
\usepackage{multirow}
\usepackage{newtxtext, newtxmath}
\usepackage{tikz}
\usepackage{diagbox}

\usepackage[noend]{algpseudocode}
\usepackage[linesnumbered,vlined,ruled,commentsnumbered]{algorithm2e}

\usepackage[capitalize]{cleveref}
\crefname{section}{Sec.}{Secs.}
\Crefname{section}{Section}{Sections}
\Crefname{table}{Table}{Tables}
\crefname{table}{Tab.}{Tabs.}

\newcommand{\circled}[1]{%
\tikz[baseline=(char.base), scale=0.6]{
\node[shape=circle,draw,inner sep=0.6pt, font=\small] (char) {#1};}}

\def\paperID{xxxx} 
\def\confName{CVPR}
\def\confYear{2026}

\title{Defending Unauthorized Model Merging via Dual-Stage Weight Protection}

\author{
Wei-Jia Chen\textsuperscript{1} \quad
Min-Yan Tsai\textsuperscript{1} \quad
Cheng-Yi Lee\textsuperscript{2} \quad
Chia-Mu Yu\textsuperscript{1}\thanks{Corresponding author.} \\
\textsuperscript{1} National Yang Ming Chiao Tung University \quad
\textsuperscript{2} Academia Sinica \\
{\tt\small \{wer9u623, chengyi.lee.1224, chiamuyu\}@gmail.com}
}

\begin{document}
\maketitle
\begin{abstract}
The rapid proliferation of pretrained models and open repositories has made model merging a convenient yet risky practice, allowing free-riders to combine fine-tuned models into a new multi-capability model without authorization. Such unauthorized model merging not only violates intellectual property rights but also undermines model ownership and accountability.
To address this issue, we present \textsc{MergeGuard}, a proactive dual-stage weight protection framework that disrupts merging compatibility while maintaining task fidelity. In the first stage, we redistribute task-relevant information across layers via $L_2$-regularized optimization, ensuring that important gradients are evenly dispersed. In the second stage, we inject structured perturbations to misalign task subspaces, breaking curvature compatibility in the loss landscape. Together, these stages reshape the model’s parameter geometry such that merged models collapse into destructive interference while the protected model remains fully functional.
Extensive experiments on both vision (\verb"ViT-L-14") and language (\verb"Llama2", \verb"Gemma2", \verb"Mistral") models demonstrate that \textsc{MergeGuard} reduces merged model accuracy by up to 90\% with less than 1.5\% performance loss on the protected model. 
\end{abstract}
\section{Introduction}\label{sec:intro}
The pretrain–finetune paradigm has become a cornerstone of modern AI, where large pretrained models are efficiently adapted to downstream domains through lightweight finetuning. This workflow, widely adopted in vision~\cite{dosovitskiy2020image}, language~\cite{brown2020language,touvron2023llama}, and multimodal~\cite{radford2021learning} communities, has been further catalyzed by open repositories such as Hugging Face~\cite{wolf2019huggingface}, GitHub~\cite{github2025}, and ModelScope~\cite{modelscope2025}, which host thousands of publicly available checkpoints. Developers can now easily obtain and recombine high-quality models without additional training. Among these techniques, model merging~\cite{yang2024model}, which integrates multiple finetuned models derived from the same pretrained source via parameter-level composition, has become a practical means to construct multi-task or cross-domain models~\cite{yang2024representation}.

\begin{figure}[!t]
  \centering
  \includegraphics[width=\columnwidth]{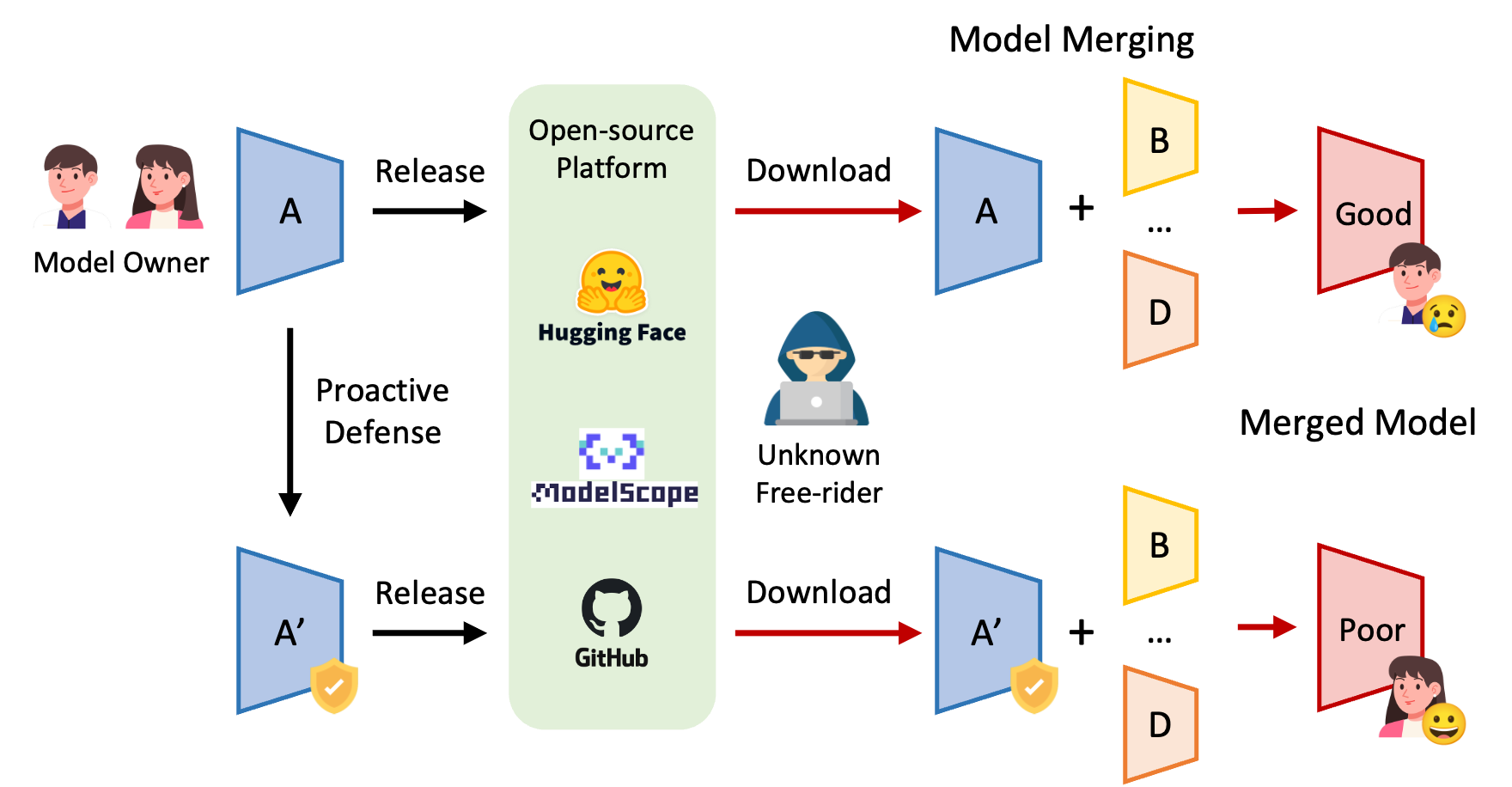}
  \caption{Unauthorized model merging and its defense.}
  \vspace{-0.15in}
  \label{fig:scenario}
\end{figure}

However, this openness also introduces a growing intellectual property risk~\cite{cong2023have,yamabemergeprint}. Free-riders can download fine-tuned models released under specific license terms and merge them to create new multi-capability models for redistribution or commercial use without authorization. Because parameter blending inherently conceals the provenance of individual weights, such \textit{unauthorized model merging}~\cite{junhao2025disrupting} allows the resulting model to inherit specialized capabilities without reproducing the original training cost or respecting licensing restrictions, posing significant challenges to model ownership~\cite{wang2022nontransferable} and attribution~\cite{li2023survey}.

To mitigate this threat, we propose a \textit{proactive} defense that preprocesses the model's parameters so that any subsequent merging will degrade functionality while preserving the model's own task performance. Designing such a defense, however, faces two key challenges. First, the defender can only modify the parameters of their own model and cannot anticipate the attacker's model, merging strategy, or target tasks, making the defense highly uncertain. Second, the defender must simultaneously maintain the model's original accuracy and suppress the merged model's effectiveness under the constraint of controlling only their own parameters (see Figure~\ref{fig:scenario}).

We make an observation that recent advances~\cite{wortsman2022model,ilharcoediting,yadav2023ties,yu2024language} in model merging often rely on sparsification~\cite{yang2025impart,qicabs}, which aims to distribute task-specific information across different layers and weights before merging. By sparsifying each fine-tuned model, their task-relevant parameters become less overlapping in the parameter space, reducing interference when the two models are later combined. Consequently, the merged model can largely retain the capabilities of each constituent model. 

Building on the above observation, we introduce a defensive two-stage preprocessing strategy, \textsc{MergeGuard}, to disrupt unauthorized model merging. The key idea is to reshape the model's weight distribution so that merging no longer preserves task performance. Specifically, we first \textit{redistribute} the parameter values, spreading those originally concentrated in a few high-magnitude weights across a wider set of parameters. This dispersal maintains the model's functionality on its original task but makes the merging process unstable, as the diffused weights are more easily amplified, diluted, or interfered with during parameter blending. However, simple averaging alone cannot fully resist merging. To further reinforce protection, we selectively adjust influential parameters by applying controlled perturbations. Because task-relevant knowledge has already been dispersed across many smaller weights, this targeted manipulation minimally affects the standalone model but substantially degrades the performance of any merged version. Through this dual adjustment, \textsc{MergeGuard} effectively preserves the integrity of the protected model while inducing collapse in unauthorized merged models.

To validate the robustness of \textsc{MergeGuard}, we consider two adaptive attacks. The first attempts to neutralize the defense by subtracting a scaled copy of the protected model's parameters from the merged weights. The second aims to estimate the hidden disturbance vector and project each gradient update orthogonally to it, avoiding movement along the defended direction during retraining. However, both strategies fail to undermine our defense (see Section~\ref{sec: Adaptive Attack}). 

\textsc{MergeGuard} has shown high effectiveness on a variety of tasks and architectures. For \verb"ViT-L-14" on the GTSRB and MNIST, the accuracy of the preprocessed model only decreased slightly (less than $1$\%). However, after merging, the accuracy of GTSRB plummeted from $86.78$\% to $12.19$\%, and MNIST from $96.02$\% to $11.35$\%. For large language models (LLMs), the accuracy after merging generally dropped to single digits, with some even falling to $0$\%. For example, the accuracy of the \verb"Gemma2" decreased from $69.6$\% to $1.52$\% on the GSM8K after merging, and from $64.02$\% to $21.34$\% on the HumanEval. 

\textbf{Contributions.} Our contributions are shown below:
\begin{itemize}
    \item We propose \textsc{MergeGuard}, a proactive defense that redistributes and perturbs model weights to prevent unauthorized merging while preserving task performance.

    \item Experiments on \verb|ViT-L-14|, \verb|Llama2|, \verb|Gemma2|, and \verb|Mistral| show that \textsc{MergeGuard} reduces merged-model accuracy by up to $90$\% with less than $1.5$\% loss on the protected model.

    \item We evaluate two adaptive attacks and show that both fail to circumvent the defense due to dispersed task information and unobservable perturbations.
\end{itemize}

\section{Related Works}
\noindent \textbf{Model Merging.}
Model merging is a training-free method for combining multiple fine-tuned models into a single multi-task model by directly manipulating their parameters. Through techniques such as weight averaging (WA)~\cite{wortsman2022model}, the parameters of several models can be linearly combined to produce a new model capable of handling multiple tasks simultaneously. Task arithmetic (TA)~\cite{ilharcoediting} further interprets tasks as vectors in the parameter space, enabling addition or subtraction between task representations. Other approaches, including TIES~\cite{chen2021evaluating} and DARE~\cite{yu2024language}, enhance merging quality via parameter sparsity and sign consistency, achieving better interpretability without additional data. More recent studies~\cite{deep2024della, choi2024revisiting, du2024parameter, lu2024twin,yang2024adamerging} extend these ideas to broader architectures and scenarios, developing increasingly flexible and efficient merging strategies.

\noindent \textbf{Defenses against Unauthorized Model Merging.}\label{sec: Defenses against Unauthorized Model Merging}
To the best of our knowledge, PaRaMS~\cite{junhao2025disrupting} is the only work that proactively defends against unauthorized model merging. In particular, PaRaMS introduces parameter rearrangement and random multi-head scaling: the former reorders MLP parameters, and the latter randomly scales attention heads while maintaining functional equivalence, thereby pushing the model out of the shared parameter basin and significantly degrading the merged model's performance by slightly harming the original task accuracy.
\section{Background Knowledge on Model Merging}\label{sec: Background Knowledge on Model Merging}
Let $\theta_{\mathrm{pre}}$ denote the parameters of the pre-trained model, and $\theta_i$ the parameters of a fine-tuned model on task $\mathcal{D}_i$. For $n$ task-specific models ${\theta_1, \dots, \theta_n}$ derived from the same pre-trained model $\theta_{\mathrm{pre}}$, model merging can be formulated as $\theta_{\mathrm{merge}} = \mathrm{Merge}(\theta_{\mathrm{pre}}, \theta_1, \dots, \theta_n)$, where Merge represents a specific parameter-level fusion strategy.

\noindent \textbf{Task Arithmetic (TA).} TA~\cite{ilharcoediting} assumes that the task-specific update from $\theta_{\mathrm{pre}}$ to $\theta_i$ encodes the distinctive capability of task $\mathcal{D}_i$. This update, referred to as the \textit{task vector} $\tau_i = \theta_i - \theta_{\mathrm{pre}}$, can be linearly combined across tasks to obtain $\theta_{\mathrm{merge}} = \theta_{\mathrm{pre}} + \lambda \sum_i \tau_i$, where $\lambda$ controls the merging scale. TA therefore constructs a multi-task model by directly superposing these task vectors.

\noindent \textbf{TIES-Merging.} Extending TA, TIES-merging~\cite{yadav2023ties} aims to alleviate interference between tasks caused by redundant or conflicting parameters. It first trims each task vector $\tau_i$ by retaining only the top-$k\%$ parameters with the largest magnitudes, then elects a unified parameter sign across tasks, and finally performs a disjoint merge that averages only the parameters consistent with the selected sign. This three-step procedure improves stability and reduces destructive interference in the merged model.

\noindent \textbf{AdaMerging.} AdaMerging~\cite{yang2024adamerging} introduces an adaptive and unsupervised model merging framework that automatically learns task-wise or layer-wise coefficients without using original training data. By minimizing prediction entropy on unlabeled samples, it adjusts merging weights to balance task contributions, thereby improving multi-task performance, generalization, and robustness. This adaptive strategy can be integrated with TA or TIES-merging to further enhance merging quality and stability.

\section{System Model}
\noindent \textbf{Attack Scenario.} 
We consider a scenario where a \textit{defender} fine-tunes a large pretrained model $\theta_{\mathrm{pre}}$ on a domain-specific dataset and releases the resulting model $\theta_{\mathrm{def}}$ to the community. A \textit{free-rider} may download this open-sourced model and merge it with their own model $\theta_{\mathrm{fr}}$, which is assumed to be derived from the same $\theta_{\mathrm{pre}}$. Critically, we focus on full-parameter fine-tuning and exclude Parameter-Efficient Fine-Tuning (PEFT) methods (\textit{e.g.}, LoRA or shallow tuning) from our threat model, as they do not expose full task vectors in the base parameter space and are thus incompatible with classical parameter-level merging. By merging the two models, the free-rider seeks to inherit the defender's specialized capabilities at minimal cost and combine them with their own model's functionality. The defender's goal is therefore to proactively modify its model $\theta_{\mathrm{def}}$ so that it maintains high performance on the intended task but degrades when merged with any other model.

\noindent \textbf{Free-rider's Capability.}
The free-rider is assumed to have limited computational resources, making expensive retraining or knowledge distillation~\cite{hinton2015distilling,wang2023survey} infeasible. In particular, because the free-rider is highly cost-sensitive, they will not perform knowledge distillation on $T_{\mathrm{def}}$ even though the dataset is downloadable, as such retraining still incurs substantially higher computation and tuning costs than direct parameter merging. Instead, they rely on low-cost model merging as a shortcut to acquire the defender's expertise. The free-rider has full white-box access to the defender's released model and a task-specific model $\theta_{\mathrm{fr}}$ derived from the same $\theta_{\mathrm{pre}}$. The attack is considered successful if the merged model $\theta_{\mathrm{merge}}$ achieves accuracy comparable to $\theta_{\mathrm{def}}$ on the defender's task $T_{\mathrm{def}}$ while preserving its own performance on $T_{\mathrm{fr}}$, where $T_X$ denotes $X$'s domain-specific task.

\noindent \textbf{Defender's Capability.}
The defender has full access to and control over the parameters of their own model $\theta_{\mathrm{def}}$ and can apply parameter-level transformations to strengthen protection. These modifications must maintain the model's original task performance while preventing the merged model from retaining it. Although the defender cannot access the free-rider's model, it can reasonably assume both originate from the same model $\theta_{\mathrm{pre}}$; otherwise, merging would not be feasible~\cite{wortsman2022model,ilharcoediting,deep2024della, choi2024revisiting, du2024parameter, lu2024twin,yang2024adamerging}. A successful defense ensures that the released model remains fully functional for legitimate use, but any merged version experiences a substantial accuracy drop on $T_{\mathrm{def}}$.

\noindent \textbf{Problem Setup.}  
The free-rider's goal is to construct a merged model  
$\theta_{\mathrm{merge}} = \text{Merge}(\theta_{\mathrm{pre}}, \theta_{\mathrm{def}}, \theta_{\mathrm{fr}})$ that simultaneously satisfies $\mathrm{Perf}(\theta_{\mathrm{merge}}, T_{\mathrm{def}}) \approx \mathrm{Perf}(\theta_{\mathrm{def}}, T_{\mathrm{def}})$ and $\mathrm{Perf}(\theta_{\mathrm{merge}}, T_{\mathrm{fr}}) \approx \mathrm{Perf}(\theta_{\mathrm{fr}}, T_{\mathrm{fr}})$, where $\mathrm{Merge}(\cdot)$ denotes a parameter-level fusion method such as model soups or task arithmetic~\cite{wortsman2022model,ilharcoediting,deep2024della,choi2024revisiting,du2024parameter,lu2024twin,yang2024adamerging}, and $\mathrm{Perf}(\cdot)$ represents the appropriate utility metric for the target model family (\textit{e.g.}, accuracy for vision or perplexity for language models). To counter this, the defender instead releases a protected model $\hat{\theta}_{\mathrm{def}} = \mathrm{Defense}(\theta_{\mathrm{def}})$ whose standalone performance remains intact, $\mathrm{Perf}(\hat{\theta}_{\mathrm{def}}, T_{\mathrm{def}}) \approx \mathrm{Perf}(\theta_{\mathrm{def}}, T_{\mathrm{def}})$, but any merged version $\hat{\theta}_{\mathrm{merge}} = \text{Merge}(\theta_{\mathrm{pre}}, \hat{\theta}_{\mathrm{def}}, \theta_{\mathrm{fr}})$ exhibits a pronounced degradation, $\mathrm{Perf}(\hat{\theta}_{\mathrm{merge}}, T_{\mathrm{def}}) \ll \mathrm{Perf}(\theta_{\mathrm{def}}, T_{\mathrm{def}})$. Therefore, our objective is to design a proactive defense function $\mathrm{Defense}(\cdot)$ that enforces this degradation property while maintaining the protected model's utility on its intended task.
\vspace{-0.02in}
\section{Proposed Method}\label{sec: Proposed Method}
To address unauthorized merging, we propose \textsc{MergeGuard}, which comprises two stages: \circled{1} Density-Aware Finetuning (training stage) and \circled{2} Adversarial Weight Negation (training-free stage). We first outline the key idea behind \textsc{MergeGuard}, then describe each stage in detail.

\vspace{0.1in}
\noindent \textbf{Key Idea.}
As discussed in Section~\ref{sec: Background Knowledge on Model Merging}, most recent model merging techniques, such as TIES~\cite{yadav2023ties} and AdaMerging~\cite{yang2024adamerging} rely on sparsification to distribute task-relevant parameters sparsely across layers, reducing interference and enabling merged models to preserve multiple capabilities. 

\begin{figure*}[!t]
  \centering
  \includegraphics[width=0.68\linewidth]{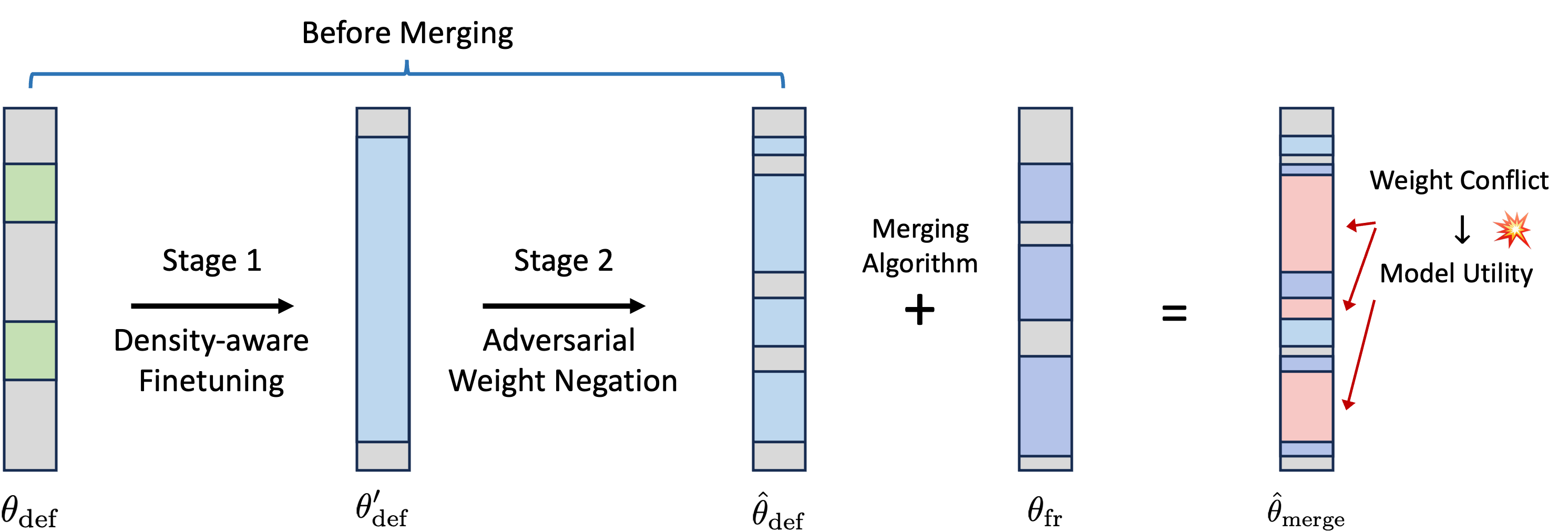}
  \caption{The workflow of \textsc{MergeGuard}.}
  \vspace{-0.1in}
  \label{fig:dense_to_sparse}
\end{figure*}

As shown in Figure~\ref{fig:dense_to_sparse}, \textsc{MergeGuard} builds on this observation from a defensive perspective. In the first stage, \textit{Density-Aware Finetuning} disperses task-critical weights through $L_2$-regularized training, ensuring that important information is evenly spread across layers rather than concentrated in a few parameters. This dispersed structure provides stability for the second stage, \textit{Adversarial Weight Negation}, which selectively offsets a subset of task-relevant directions to intentionally misalign them with potential merging counterparts. Because the weight importance has already been evenly distributed by \textit{Stage 1}, these perturbations preserve the model's own task performance while disrupting the cross-model compatibility required for successful merging. Together, the two stages reshape the parameter landscape so that merging attempts result in destructive interference, effectively neutralizing unauthorized model combination.

\vspace{0.1in}
\noindent \textbf{Stage 1: Density-Aware Finetuning.}
The goal of this stage is to disperse task-critical weights across different layers through training. 
In addition to the standard cross-entropy loss $L_{\mathrm{CE}}$ used in classification, we introduce an $L_2$ regularization term that drives model parameters toward zero, encouraging a more balanced distribution of important information. The total loss function is formulated as:
\begin{equation}\label{eq: 1}
    L_{\mathrm{Total}} = L_{\mathrm{CE}} + \alpha L_2
\end{equation}
Specifically, for vision tasks, the loss can be explicitly expanded as:
{\setlength{\abovedisplayskip}{4pt}
\setlength{\belowdisplayskip}{4pt}
\begin{equation}\label{eq: 2}
    L_{\mathrm{Total}} = -\sum_{t=1}^{T} y_t \log(\hat{y}_t) + \alpha \sum_{\ell=1}^{L} \|\theta^{(\ell)}\|_2^2,
\end{equation}}
where $y_t$ and $\hat{y}_t$ denote the ground-truth label and the predicted probability for the $t$-th class, respectively; $\alpha$ is a hyperparameter balancing the two terms; and $\theta^{(\ell)}$ represents the trainable parameters of the $\ell$-th layer. Crucially, the $L_2$ penalty is applied layer-wise so that each layer is regularized independently.

Similarly, for LLMs, the objective function is formulated based on next-token prediction:
{\setlength{\abovedisplayskip}{4pt}
\setlength{\belowdisplayskip}{4pt}
\begin{equation}
    L_{\mathrm{Total}} = -\sum_{t=1}^{T} \log P(y_t \mid y_{<t}, x; \theta) + \alpha \sum_{\ell=1}^{L} \|\theta^{(\ell)}\|_2^2,
\end{equation}}
where $P(y_t \mid y_{<t}, x; \theta)$ denotes the conditional probability of predicting the token $y_t$ given the preceding context $y_{<t}$ and input $x$. This formulation applies $L_2$ regularization across all layer parameters, effectively smoothing large-magnitude weights and promoting an even distribution of task-relevant signals throughout the network layers.


Overall, given $\theta_{\mathrm{def}}$, we obtain $\theta'_{\mathrm{def}}$ after \textit{Stage 1}. By reducing the concentration of dominant weights, this stage effectively disperses important information throughout the model, making subsequent merging operations less stable and thereby diminishing the merged model's performance.

\vspace{0.10in}
\noindent \textbf{Stage 2: Adversarial Weight Negation.}
This stage introduces a training-free procedure that intentionally offsets task-relevant weights to disrupt merging compatibility.  
We begin by masking each layer of the model and measuring the resulting accuracy drop.  
The top $k'\%$ of layers that cause the largest performance degradation are identified as critical and excluded from subsequent operations. Afterwards, a binary masking vector $M$ is then constructed, where elements corresponding to $k'\%$ critical layers and to the least significant $(1-k)(1-k')\%$ parameters are set to zero, and those in the remaining $k(1-k')\%$ range are set to one.  
This selective preservation ensures that later weight adjustments do not damage essential functionality. Next, we calculate 
\begin{align}\label{eq: 3}
\hat{\theta}_{\mathrm{def}} = \theta'_{\mathrm{def}} - \beta \, M \odot \tau'_{\mathrm{def}},
\end{align} where $\odot$ denotes the inner product, $\tau'_{\mathrm{def}} = \theta'_{\mathrm{def}} - \theta_{\mathrm{pre}}$ denotes the task vector representing the direction of the defender's fine-tuning, and $\beta > 0$ is a small perturbation factor. Subtracting $\tau_{\mathrm{def}}$ in the masked parameter space weakens certain informative directions without compromising the core task performance. Consequently, the protected model remains stable on its own task, but any merged version suffers severe degradation due to misaligned task directions and dilution introduced by the merging algorithm.

We note that some layers differ substantially in parameter scale across tasks, and applying this operation directly can destabilize the model. To prevent collapse in LLMs, we exclude \textit{model.embed\_tokens.weight}, \textit{model.norm.weight}, and \textit{lm\_head.weight} from the subtraction process, while ViT-based models undergo the full operation.

\vspace{0.1in}
\noindent \textbf{Theoretical Justification of \textsc{MergeGuard}.}
Model merging can be expressed as a parameter-level combination:
\begin{align}
\theta_{\mathrm{merge}} = \theta_{\mathrm{pre}} + \lambda_1 (\theta_{\mathrm{def}} - \theta_{\mathrm{pre}}) + \lambda_2 (\theta_{\mathrm{fr}} - \theta_{\mathrm{pre}}),
\end{align} where $\lambda_1, \lambda_2 \in [0,1]$ control the blending ratio. A successful merge implicitly assumes that both task vectors $\tau_{\mathrm{def}} = \theta_{\mathrm{def}} - \theta_{\mathrm{pre}}$ and $\tau_{\mathrm{fr}} = \theta_{\mathrm{fr}} - \theta_{\mathrm{pre}}$ reside in a \emph{shared linear subspace} representing compatible gradient directions~\cite{wei2025modeling,tam2024merging,zhou2025taskvectorsgradients}. In this case, their superposition still corresponds to a descent direction in the loss landscape~\cite{fort2025basinlike,li2018visualizing}, preserving both tasks.

\textsc{MergeGuard} breaks this assumption by jointly modifying both the \emph{magnitude distribution} and the \emph{directional alignment} of $\tau_{\mathrm{def}}$.
Formally, after \textit{Stage 1}, the finetuned parameters become $\theta'_{\mathrm{def}} = \theta_{\mathrm{pre}} + \tau'_{\mathrm{def}}$, where $\tau'_{\mathrm{def}} = \tau_{\mathrm{def}} + \epsilon$, $\mathbb{E}[\epsilon] = 0$, and $\text{Cov}[\epsilon] = \sigma^2 \textbf{I}$. This step redistributes task-relevant gradients more isotropically, flattening the curvature of the local basin and ensuring that important information is evenly spread rather than concentrated along a few dominant directions. As a result, the Hessian eigenvectors associated with large eigenvalues that previously spanned the shared merging subspace become decorrelated.

Then, \textit{Stage 2} injects a small but structured perturbation along a \emph{locally adversarial} direction: $\widehat{\theta}_{\mathrm{def}} = \theta'_{\mathrm{def}} - \beta \, M \odot \tau'_{\mathrm{def}}$ (Eq.~\ref{eq: 3}). This operation effectively rotates the task vector by an angle $\phi = \arccos \frac{\langle \tau'_{\mathrm{def}}, \tau_{\mathrm{fr}}\rangle}{\|\tau'_{\mathrm{def}}\|\|\tau_{\mathrm{fr}}\|}$, thereby increasing the expected interference term in the merged loss:
\begin{align}
\Delta \mathcal{L}_{\mathrm{merge}} \approx \lambda_1 \lambda_2 \, \|\tau'_{\mathrm{def}}\| \, \|\tau_{\mathrm{fr}}\| \, (1 - \cos\phi),
\end{align} which grows quadratically with the misalignment angle.
In practice, even small rotations (e.g., $\phi>30^\circ$) suffice to push the merged model outside the shared basin, causing the loss landscape to exhibit destructive interference between the two task vectors. As the masked perturbation preserves low-order statistics within each critical layer, the standalone model maintains its original functionality, while any linear or sparsity-based fusion~\cite{wortsman2022model,ilharcoediting,deep2024della,choi2024revisiting,du2024parameter,lu2024twin,yang2024adamerging, yu2024language} collapses due to incompatible curvature directions.

Hence, \textsc{MergeGuard}'s defense can be theoretically interpreted as \emph{inducing curvature misalignment} in the task subspace, transforming the previously compatible manifold of fine-tuned models into disjoint basins. This ensures that the merged parameter $\theta_{\mathrm{merge}}$ no longer lies near a stationary point of either task's loss function, leading to substantial degradation in unauthorized merged models.
\section{Evaluation} \label{sec6:expt}

\subsection{Experiment Setups}
\noindent \textbf{Models.} 
For the image classification task, we adopt the \verb"ViT-L-14"~\cite{dosovitskiy2020image} architecture, a transformer-based vision backbone widely used for downstream adaptation. 
For image generation, we employ Stable Diffusion 1.5 (SD1.5)~\cite{rombach2022high}, a latent diffusion model trained on large-scale text–image pairs. For the language tasks, we evaluate three representative architectures: \verb"Llama2-7B"~\cite{touvron2023llama}, \verb"Gemma2-9B"~\cite{team2024gemma}, and \verb"Mistral-7B"~\cite{jiang2023mistral7b}. 

\noindent \textbf{Datasets.} 
For image classification, we evaluate across eight diverse datasets: SUN397~\cite{xiao2010sun}, Cars~\cite{krause20133d}, RESISC45~\cite{cheng2017remote}, SVHN~\cite{netzer2011reading}, GTSRB~\cite{stallkamp2011german}, MNIST~\cite{lecun1998mnist}, EuroSAT~\cite{helber2019eurosat}, and DTD~\cite{cimpoi2014describing}, which span natural scenes, fine-grained objects, digits, satellite imagery, and textures. For image generation, we consider WikiArt~\cite{wikiart2025} and Naruto~\cite{cervenka2022naruto2}. For LLM evaluation, we employ AlpacaEval~\cite{li2023alpacaeval} for instruction following, GSM8K~\cite{cobbe2021training} for mathematical reasoning, and HumanEval~\cite{chen2021evaluating} for program synthesis. Details about these datasets are shown in the Appendix.


\noindent \textbf{Metrics.} 
In image classification, we report top-1 accuracy to assess recognition capability. 
For LLMs, the AlpacaEval benchmark reports the win rate against \verb"Zephyr-7B"~\cite{tunstallzephyr}; 
the GSM8K benchmark measures the proportion of correctly solved problems; 
and the HumanEval benchmark measures pass@1 accuracy~\cite{chen2021evaluating}, indicating the fraction of valid programs that pass all test cases.

\noindent \textbf{Merging Methods.} 
We assess the robustness of \textsc{MergeGuard} under multiple representative model-merging paradigms, including Weight Averaging (WA)~\cite{wortsman2022model}, Task Arithmetic (TA)~\cite{ilharcoediting}, TIES-Merging (TIES)~\cite{yadav2023ties}, and AdaMerging (ADA)~\cite{yang2024adamerging}, with DARE~\cite{yu2024language} additionally employed as an optional pre-processing module to form DT (DARE+TIES) used in LLMs. 

\noindent \textbf{Baselines.} 
Following prior work, PaRaMS~\cite{junhao2025disrupting} serves as the only proactive defense specifically designed against unauthorized model merging. 
Hence, we use PaRaMS as the sole baseline for quantitative comparison.

\noindent \textbf{Implementation of \textsc{MergeGuard}.} Our implementation of the \textit{Density-Aware Finetuning} stage requires no additional hyperparameter tuning. The \textit{Adversarial Weight Negation} stage introduces two key hyperparameters: $k'$ and $k$, which control the proportion of preserved and perturbed layers, respectively. For all experiments, we maintain a fixed hyperparameter set of $k'=10$, $k=0.1$, $\alpha=0.01$, and $\beta=1$, eliminating the need for task-specific tuning. Merging coefficients are determined following the protocols of the respective model merging methods~\cite{wortsman2022model,ilharcoediting,yadav2023ties,yang2024adamerging,yu2024language}.

\begin{table}[!ht]
\centering
\caption{Standalone model accuracy before/after \textsc{MergeGuard}.}
\vspace{-0.05in}
\resizebox{\columnwidth}{!}{
\begin{tabular}{l|cccccccc}
\toprule
 & SUN397 & Cars & RESISC45 & EuroSAT & SVHN & GTSRB & MNIST & DTD \\
\midrule
$\theta_{\mathrm{pre}}$ & 68.24 & 77.94 & 71.33 & 62.72 & 58.45 & 50.55 & 76.36 & 55.37 \\
\midrule
$\theta_{\mathrm{def}}$ & 82.32 & 92.39 & 97.37 & 99.81 & 98.11 & 99.24 & 99.69 & 84.15 \\
\midrule
$\hat{\theta}_{\mathrm{def}}$ & 81.52 & 88.29 & 97.25 & 95.46 & 96.82 & 98.25 & 99.27 & 82.16 \\
\bottomrule
\end{tabular}}
\label{table:benign}
\end{table}


\subsection{Experimental Results}
\subsubsection{Results on Image Classification}
Table~\ref{table:benign} summarizes the classification accuracy of $\theta_{\mathrm{pre}}$, $\theta_{\mathrm{def}}$, and $\hat{\theta}_{\mathrm{def}}$. As expected, fine-tuning $\theta_{\mathrm{pre}}$ on $T_{\mathrm{def}}$ substantially improves performance, confirming the effectiveness of task adaptation. Comparing $\theta_{\mathrm{def}}$ with $\hat{\theta}_{\mathrm{def}}$, we observe that \textsc{MergeGuard} maintains nearly identical accuracy, introducing only a negligible drop, which demonstrates that the protection process preserves the model's original utility.

\begin{table*}[!ht]
\centering
\caption{Classification accuracy of the merged models without \textsc{MergeGuard} ($\theta_{\mathrm{merge}}$) and with \textsc{MergeGuard} ($\hat{\theta}_{\mathrm{merge}}$).}
\vspace{-0.05in}
\resizebox{0.9\linewidth}{!}{
\begin{tabular}{l|cc|cc|cc|cc|cc|cc|cc|cc}
\toprule
\multirow{2}{*}{Method} &
\multicolumn{2}{c}{SUN397} &
\multicolumn{2}{c}{Cars} &
\multicolumn{2}{c}{RESISC45} &
\multicolumn{2}{c}{EuroSAT} &
\multicolumn{2}{c}{SVHN} &
\multicolumn{2}{c}{GTSRB} &
\multicolumn{2}{c}{MNIST} &
\multicolumn{2}{c}{DTD} \\
\cmidrule(lr){2-3}\cmidrule(lr){4-5}\cmidrule(lr){6-7}\cmidrule(lr){8-9}
\cmidrule(lr){10-11}\cmidrule(lr){12-13}\cmidrule(lr){14-15}\cmidrule(lr){16-17}

& $\theta_{\mathrm{merge}}$ & $\hat{\theta}_{\mathrm{merge}}$ & $\theta_{\mathrm{merge}}$ & $\hat{\theta}_{\mathrm{merge}}$ & $\theta_{\mathrm{merge}}$ & $\hat{\theta}_{\mathrm{merge}}$ & $\theta_{\mathrm{merge}}$ & $\hat{\theta}_{\mathrm{merge}}$ & $\theta_{\mathrm{merge}}$ & $\hat{\theta}_{\mathrm{merge}}$ & $\theta_{\mathrm{merge}}$ & $\hat{\theta}_{\mathrm{merge}}$ & $\theta_{\mathrm{merge}}$ & $\hat{\theta}_{\mathrm{merge}}$ & $\theta_{\mathrm{merge}}$ & $\hat{\theta}_{\mathrm{merge}}$ \\
\midrule
WA~\cite{wortsman2022model} & 72.1 & 56.16 & 81.6 & 76.63 & 82.6 & 4.98 & 91.9 & 13.07 & 78.2 & 10.05 & 70.7 & 44.28 & 97.1 & 38.97 & 62.8 & 2.61 \\
TA~\cite{ilharcoediting}
 & 73.9 & 54.27 & 82.1 & 38.24 & 86.6 & 56.50 & 94.1 & 54.94 & 87.9 & 47.28 & 86.7 & 12.91 & 98.9 & 11.35 & 65.6 & 46.65 \\
TIES~\cite{yadav2023ties} & 76.5 & 70.77 & 85.0 & 45.64 & 89.3 & 33.51 & 95.7 & 14.48 & 90.3 & 7.77 & 83.3 & 4.44 & 99.0 & 35.70 & 68.6 & 48.88 \\
ADA~\cite{yang2024adamerging} & 79.4 & 65.51 & 90.3 & 74.19 & 91.6 & 72.27 & 97.4 & 11.22 & 93.4 & 7.76 & 97.5 & 2.19 & 99.0 & 71.37 & 79.2 & 49.20 \\
\bottomrule
\end{tabular}}
\label{table: comparison to different merging methods}
\end{table*}

\begin{table*}[!ht]
\centering
\caption{The comparison between \textsc{MergeGuard} and PaRaMS~\cite{junhao2025disrupting}. Each number corresponds to the classification accuracy (\%). The average accuracy drop is $30.76$ for PaRaMS and is $52.11$ for \textsc{MergeGuard}.}
\vspace{-0.05in}
\resizebox{0.95\linewidth}{!}{
\begin{tabular}{l|cc|cc|cc|cc|cc|cc|cc|cc}
\toprule
\multirow{2}{*}{Methods} &
\multicolumn{2}{c}{SUN397} &
\multicolumn{2}{c}{Cars} &
\multicolumn{2}{c}{RESISC45} &
\multicolumn{2}{c}{EuroSAT} &
\multicolumn{2}{c}{SVHN} &
\multicolumn{2}{c}{GTSRB} &
\multicolumn{2}{c}{MNIST} &
\multicolumn{2}{c}{DTD} \\
\cmidrule(lr){2-3}\cmidrule(lr){4-5}\cmidrule(lr){6-7}\cmidrule(lr){8-9}
\cmidrule(lr){10-11}\cmidrule(lr){12-13}\cmidrule(lr){14-15}\cmidrule(lr){16-17}
 & $\hat{\theta}_{\mathrm{def}}$ & $\hat{\theta}_{\mathrm{merge}}$ & $\hat{\theta}_{\mathrm{def}}$ & $\hat{\theta}_{\mathrm{merge}}$ & $\hat{\theta}_{\mathrm{def}}$ & $\hat{\theta}_{\mathrm{merge}}$ & $\hat{\theta}_{\mathrm{def}}$ & $\hat{\theta}_{\mathrm{merge}}$ & $\hat{\theta}_{\mathrm{def}}$ & $\hat{\theta}_{\mathrm{merge}}$ & $\hat{\theta}_{\mathrm{def}}$ & $\hat{\theta}_{\mathrm{merge}}$ & $\hat{\theta}_{\mathrm{def}}$ & $\hat{\theta}_{\mathrm{merge}}$ & $\hat{\theta}_{\mathrm{def}}$ & $\hat{\theta}_{\mathrm{merge}}$ \\
\midrule
PaRaMS~\cite{junhao2025disrupting} & 82.32 & 56.77 & 92.39 & 38.61 & 97.37 & 67.86 & 99.81 & 74.81 & 98.11 & 71.95 & 99.24 & 52.25 & 99.69 & 96.02 & 84.15 & 48.72 \\
\textsc{MergeGuard} & 81.52 & 54.27 & 88.29 & 38.24 & 97.25 & 56.50 & 95.46 & 54.94 & 96.82 & 47.28 & 98.25 & 12.91 & 99.27 & 11.35 & 82.16 & 46.65 \\
\bottomrule
\end{tabular}}
\label{table: comparison to PaRaMS}
\end{table*}

Table~\ref{table: comparison to different merging methods} summarizes the image classification results on \textsc{MergeGuard} under different model merging methods. 
Across all evaluated merging paradigms, \textsc{MergeGuard} consistently enforces significant degradation on the merged models while preserving the original model's performance. 
In particular, the preprocessing causes only a marginal reduction (typically less than 1\%) in standalone accuracy, indicating that the dual-stage mechanism minimally affects the base model's functionality. 
However, once the protected models are merged, their accuracy drops drastically across all datasets. 
Compared to the unprotected or baseline configurations, \textsc{MergeGuard} achieves much larger post-merge degradation across all merging strategies.

Table~\ref{table: comparison to PaRaMS} summarizes the image classification results comparing \textsc{MergeGuard} with PaRaMS~\cite{junhao2025disrupting}. 
During the preprocessing stage, \textsc{MergeGuard} introduces a minor accuracy drop on the original model due to the regularization and perturbation applied in the protection process. 
In contrast, PaRaMS~\cite{junhao2025disrupting}, which only performs parameter rearrangement, preserves the original accuracy almost perfectly. 
However, after model merging, the gap between the two defenses becomes substantial. 
\textsc{MergeGuard} consistently enforces stronger degradation on the merged models while maintaining near-original accuracy on the protected model. 
For instance, on GTSRB and MNIST, \textsc{MergeGuard} in TA reduced the original accuracy by less than $2$\%, yet the merged accuracy dropped sharply from $86.78$\% to $12.91$\% and from $98.94$\% to $11.35$\%, respectively. By comparison, PaRaMS only reduced the same metrics to $52.25$\% and $96.02$\%. These results confirm that \textsc{MergeGuard} substantially weakens merging compatibility.

\begin{figure*}[!ht]
  \centering
  \includegraphics[width=0.9\linewidth]{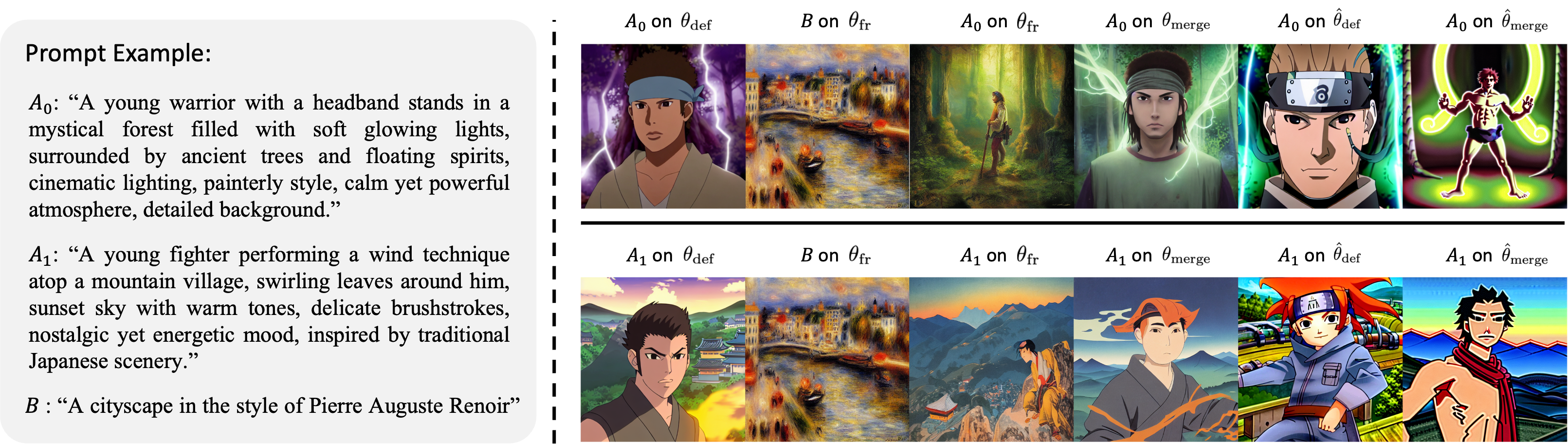}
  \vspace{-0.05in}
  \caption{Visualization Results of SD 1.5.}
  \label{fig:diffusion}
\end{figure*}

\begin{figure*}[!ht]
  \centering
  \begin{subfigure}[b]{0.23\linewidth}
    \includegraphics[width=\linewidth]{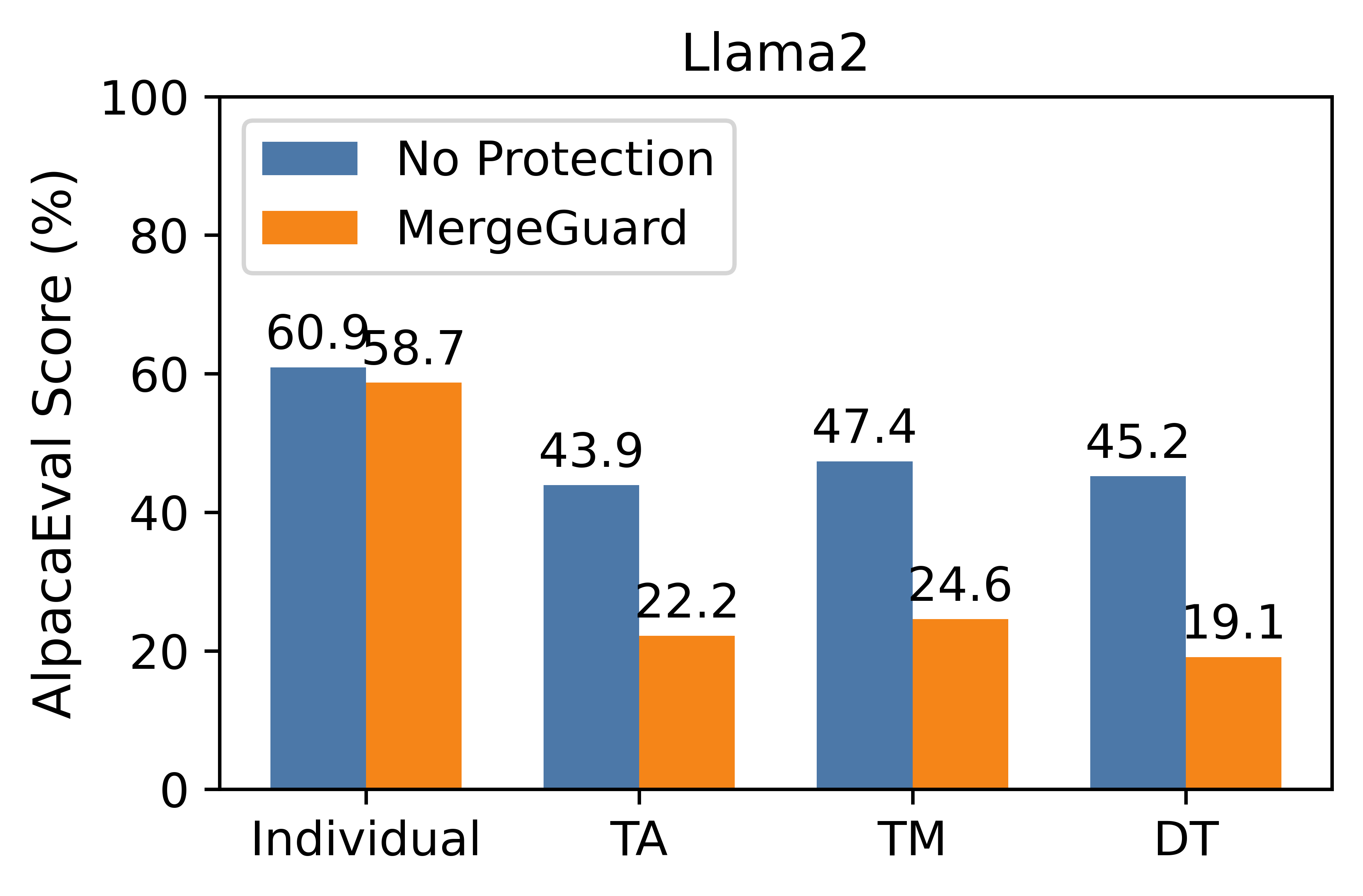}
  \end{subfigure}
  \hspace{-0.5em}
  \begin{subfigure}[b]{0.23\linewidth}
    \includegraphics[width=\linewidth]{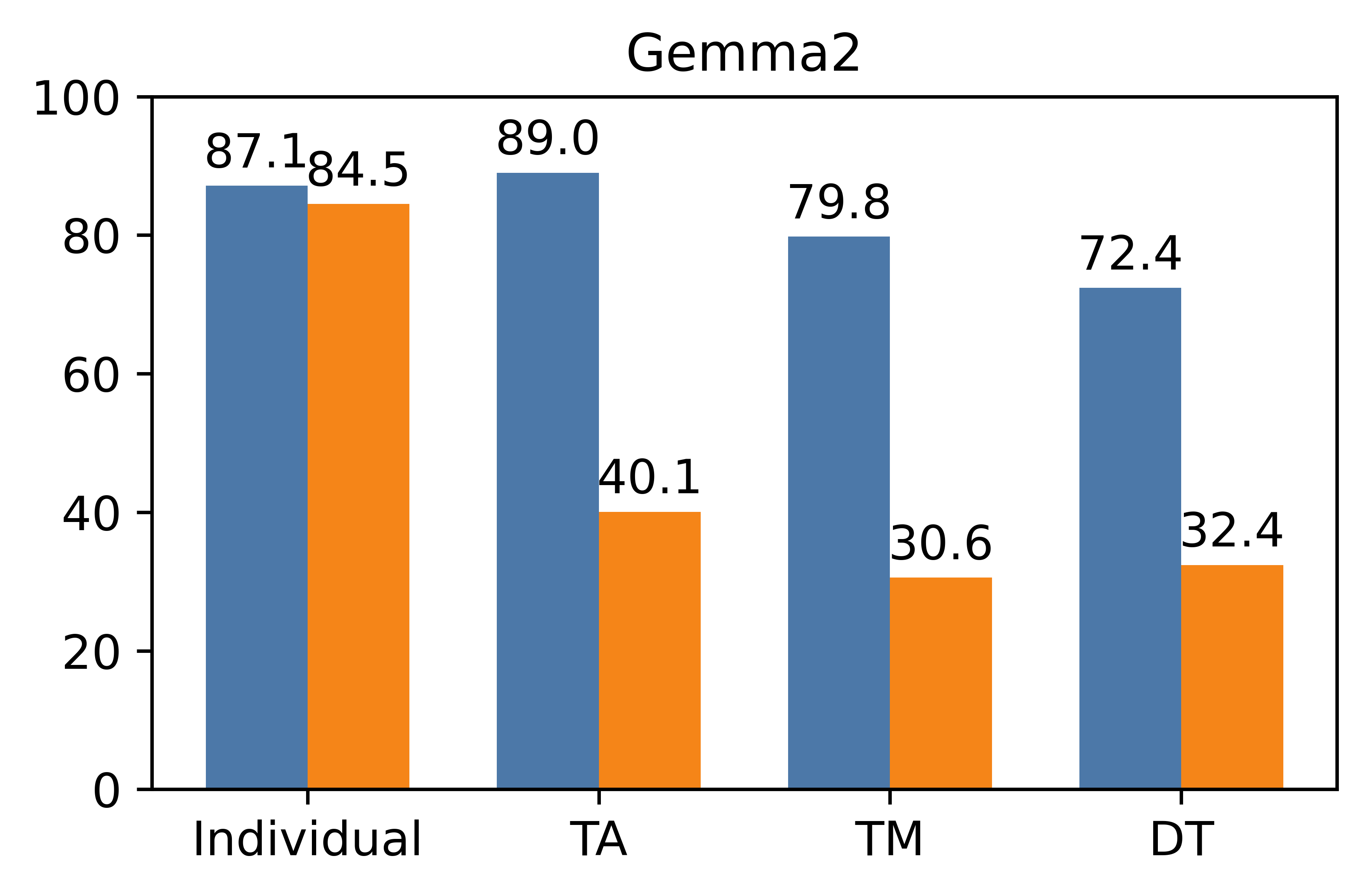}
  \end{subfigure}
  \hspace{-0.5em}
  \begin{subfigure}[b]{0.23\linewidth}
    \includegraphics[width=\linewidth]{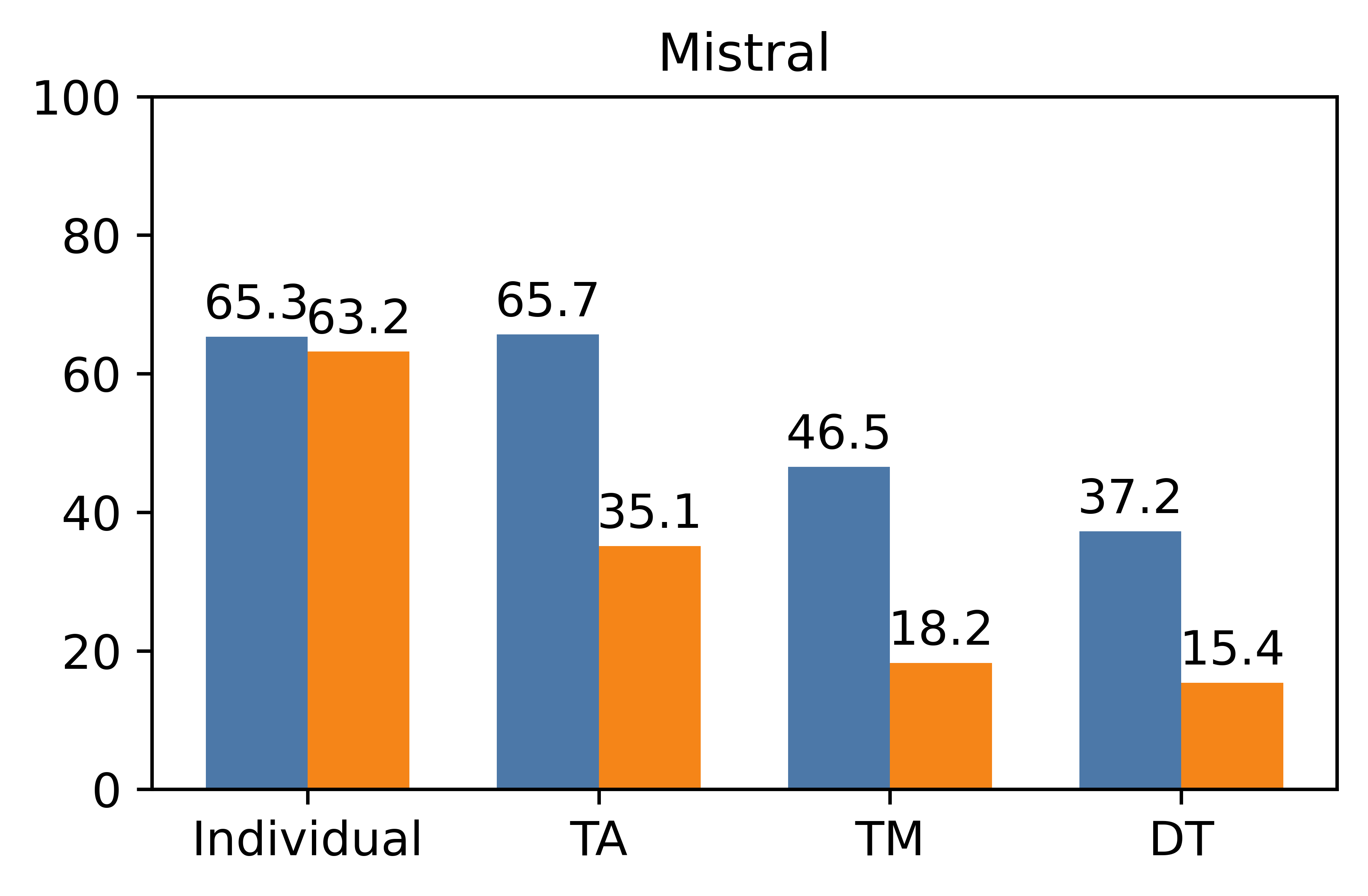}
  \end{subfigure}
  \\ \vspace{-0.3em}
  \begin{subfigure}[b]{0.23\linewidth}
    \includegraphics[width=\linewidth]{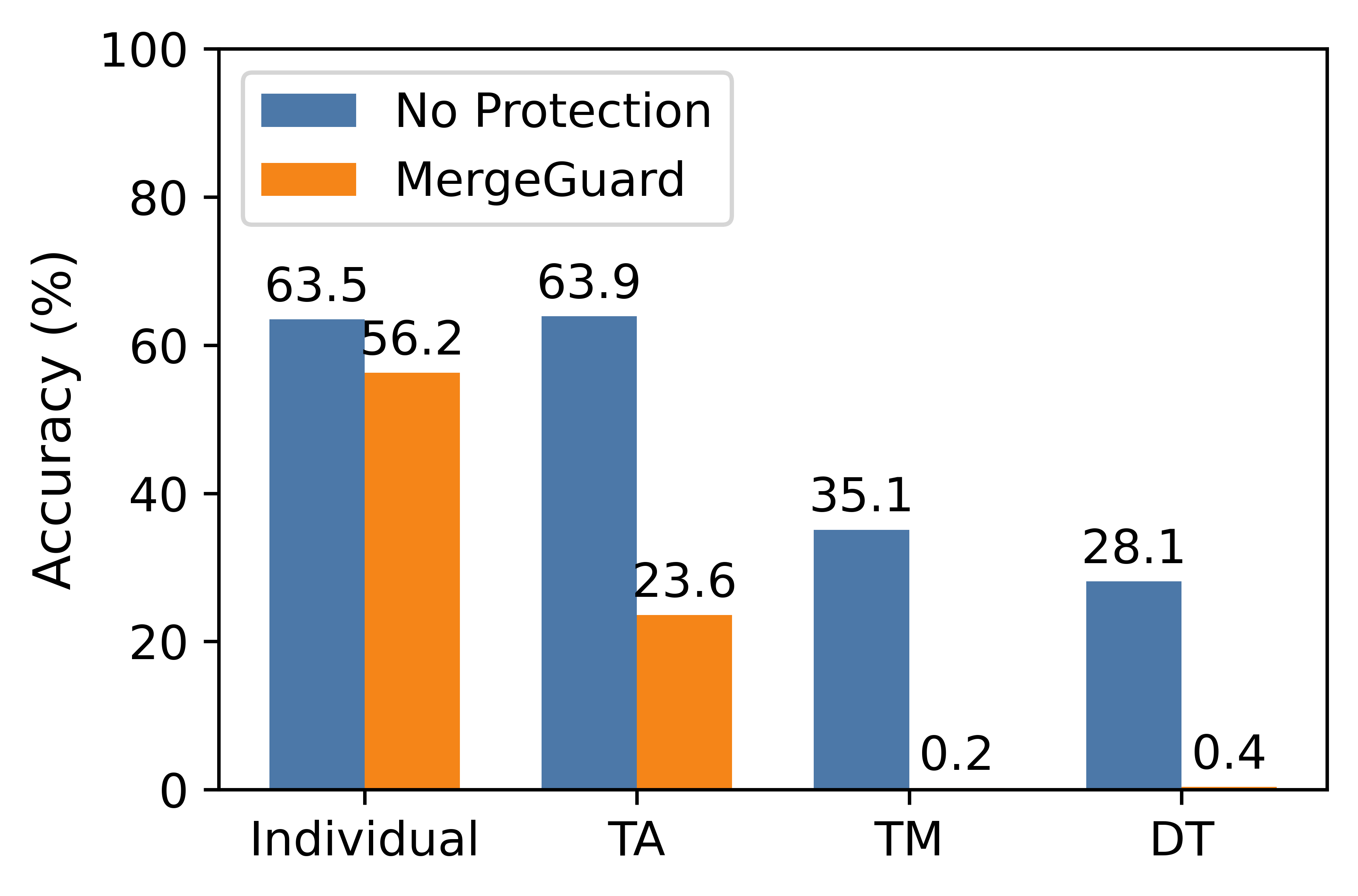}
  \end{subfigure}
  \hspace{-0.5em}
  \begin{subfigure}[b]{0.23\linewidth}
    \includegraphics[width=\linewidth]{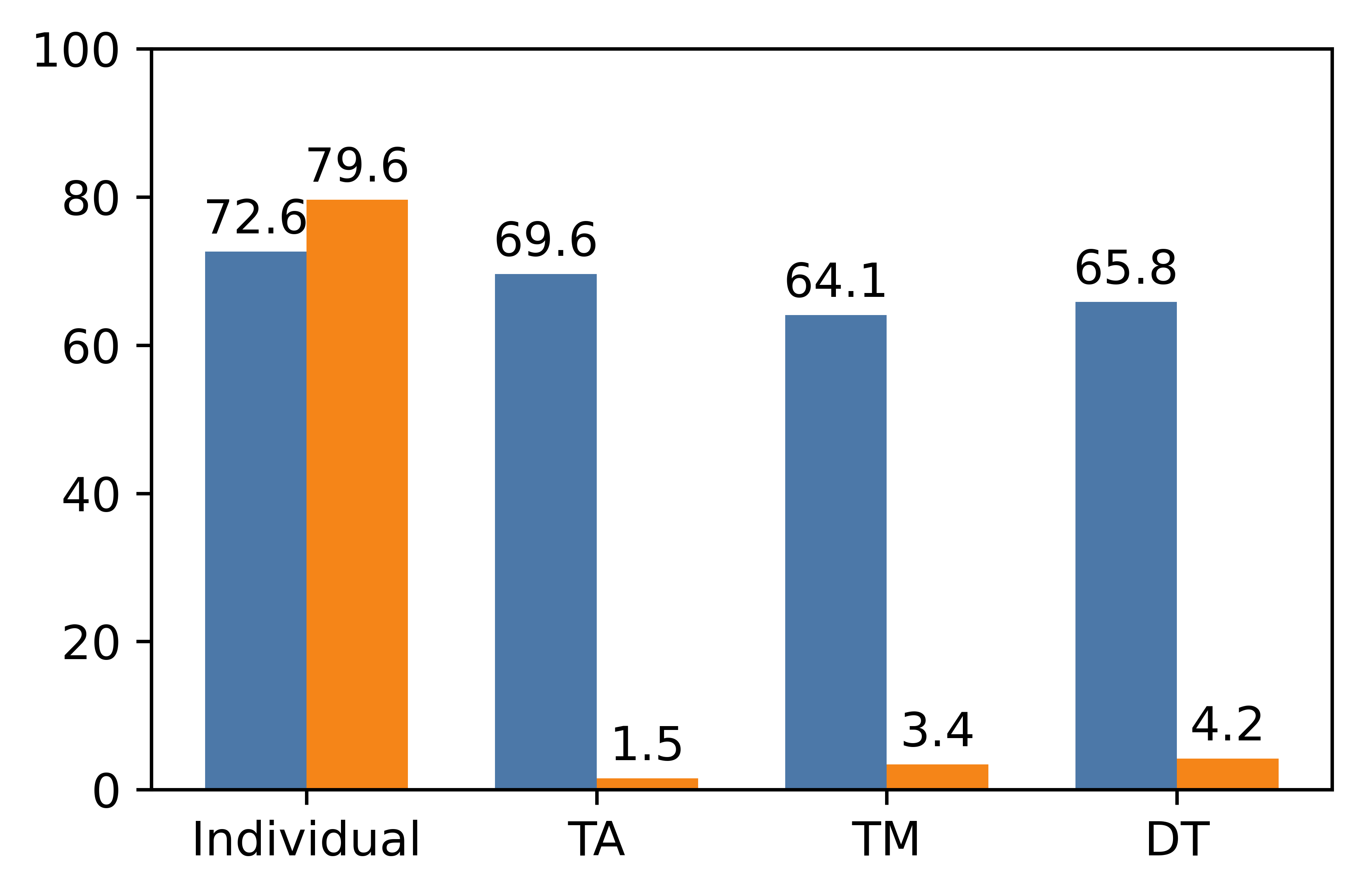}
  \end{subfigure}
  \hspace{-0.5em}
  \begin{subfigure}[b]{0.23\linewidth}
    \includegraphics[width=\linewidth]{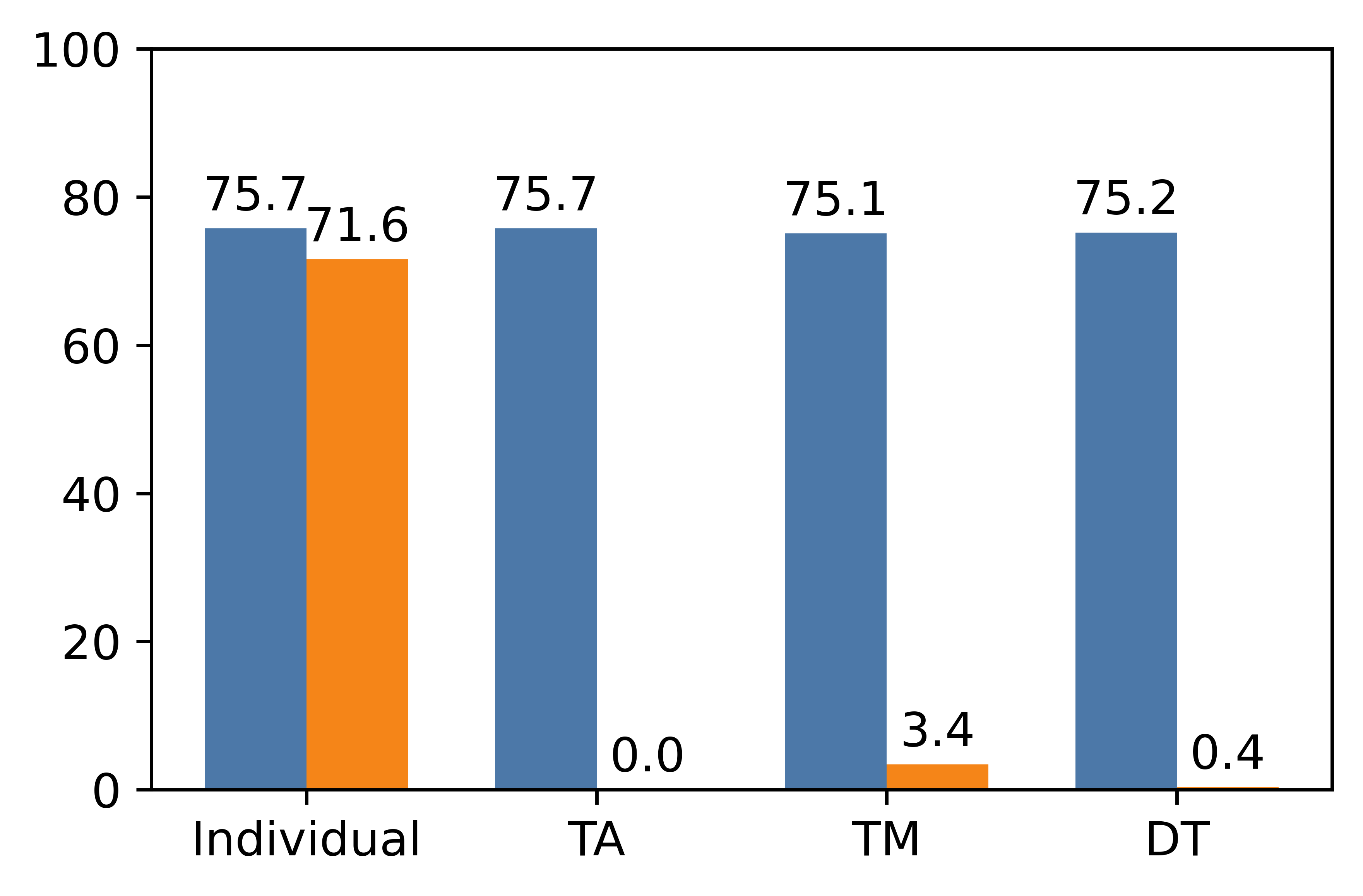}
  \end{subfigure}
  \\ \vspace{-0.3em}
  \begin{subfigure}[b]{0.23\linewidth}
    \includegraphics[width=\linewidth]{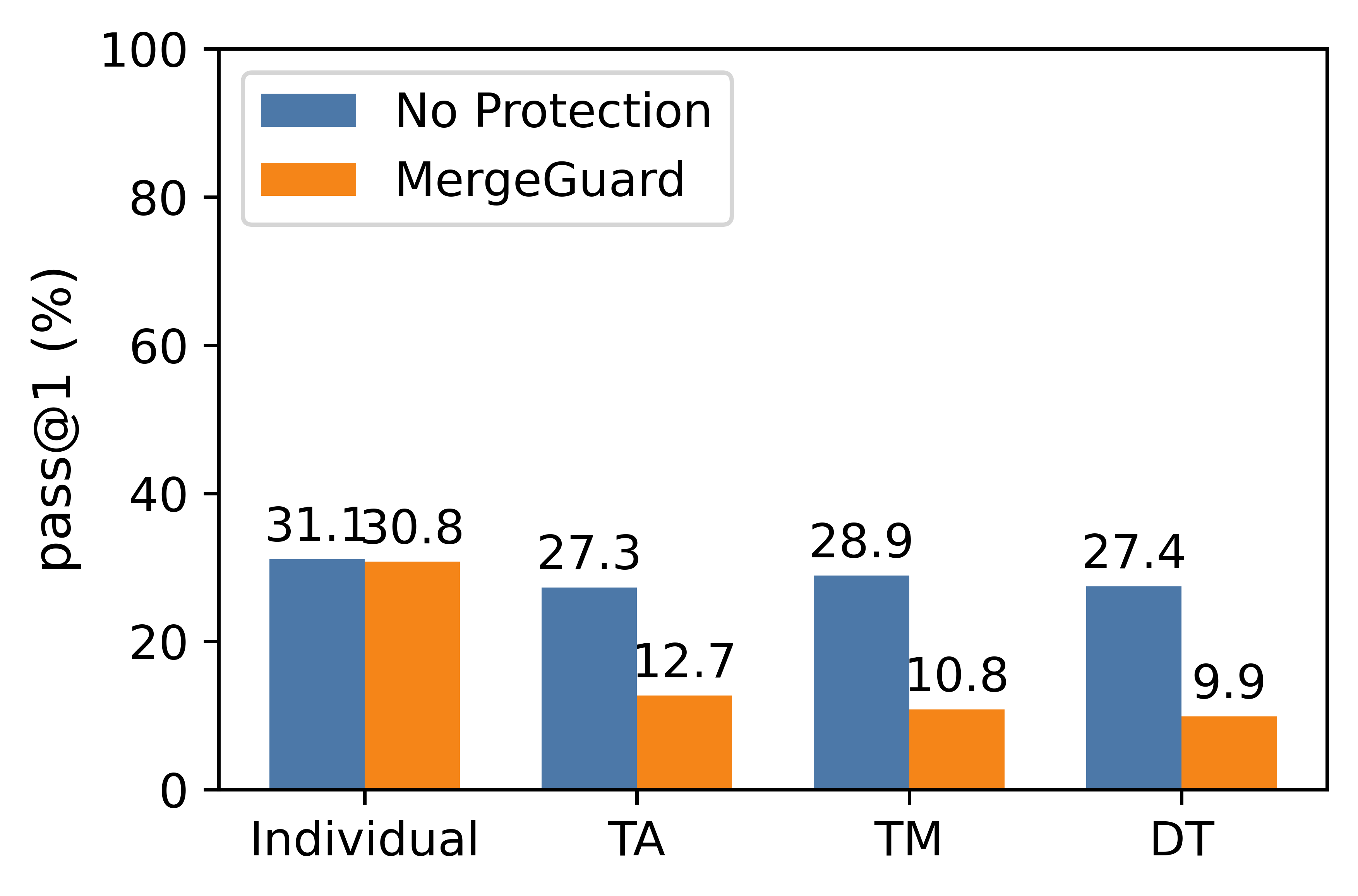}
  \end{subfigure}
  \hspace{-0.5em}
  \begin{subfigure}[b]{0.23\linewidth}
    \includegraphics[width=\linewidth]{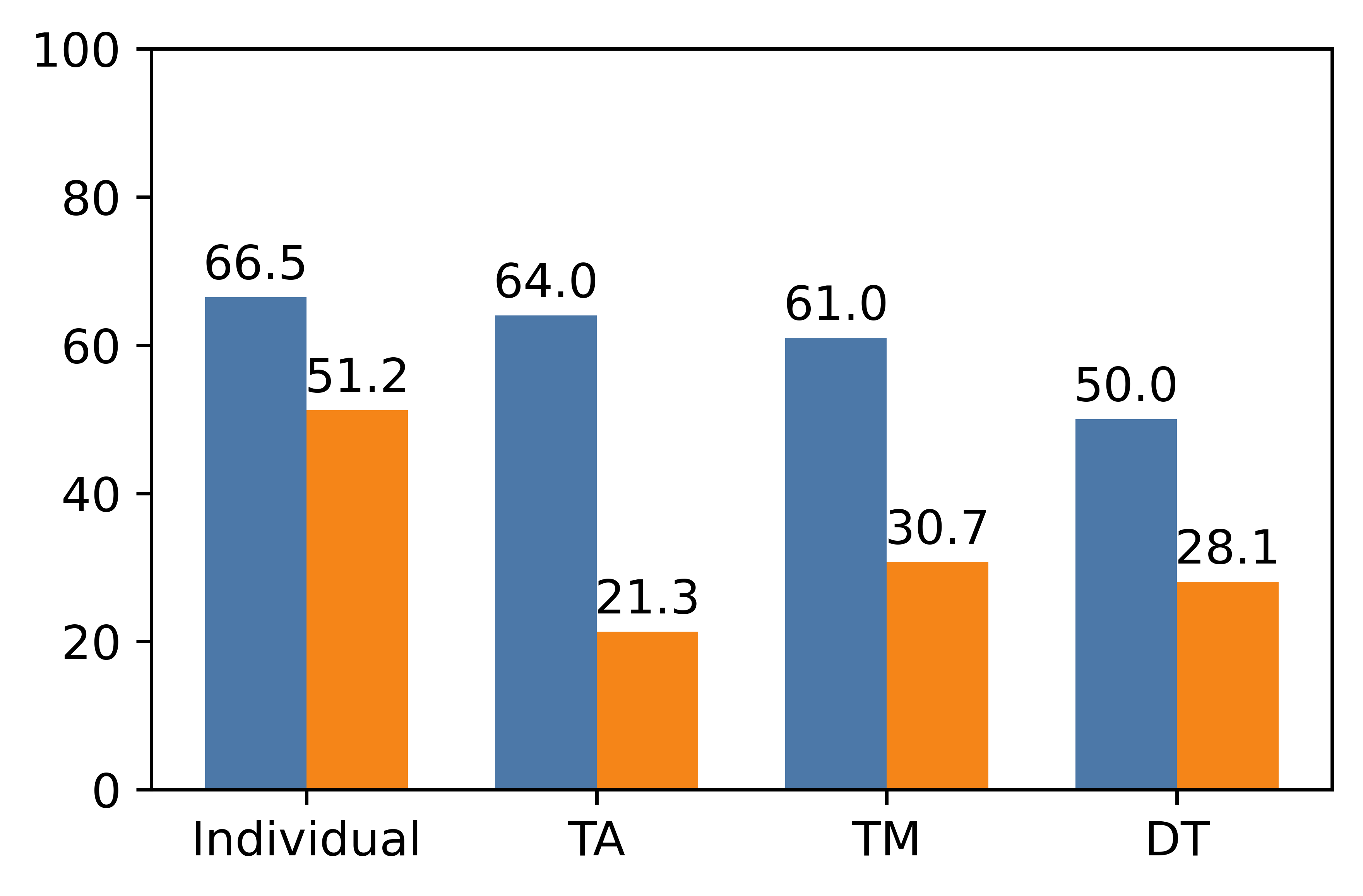}
  \end{subfigure}
  \hspace{-0.5em}
  \begin{subfigure}[b]{0.23\linewidth}
    \includegraphics[width=\linewidth]{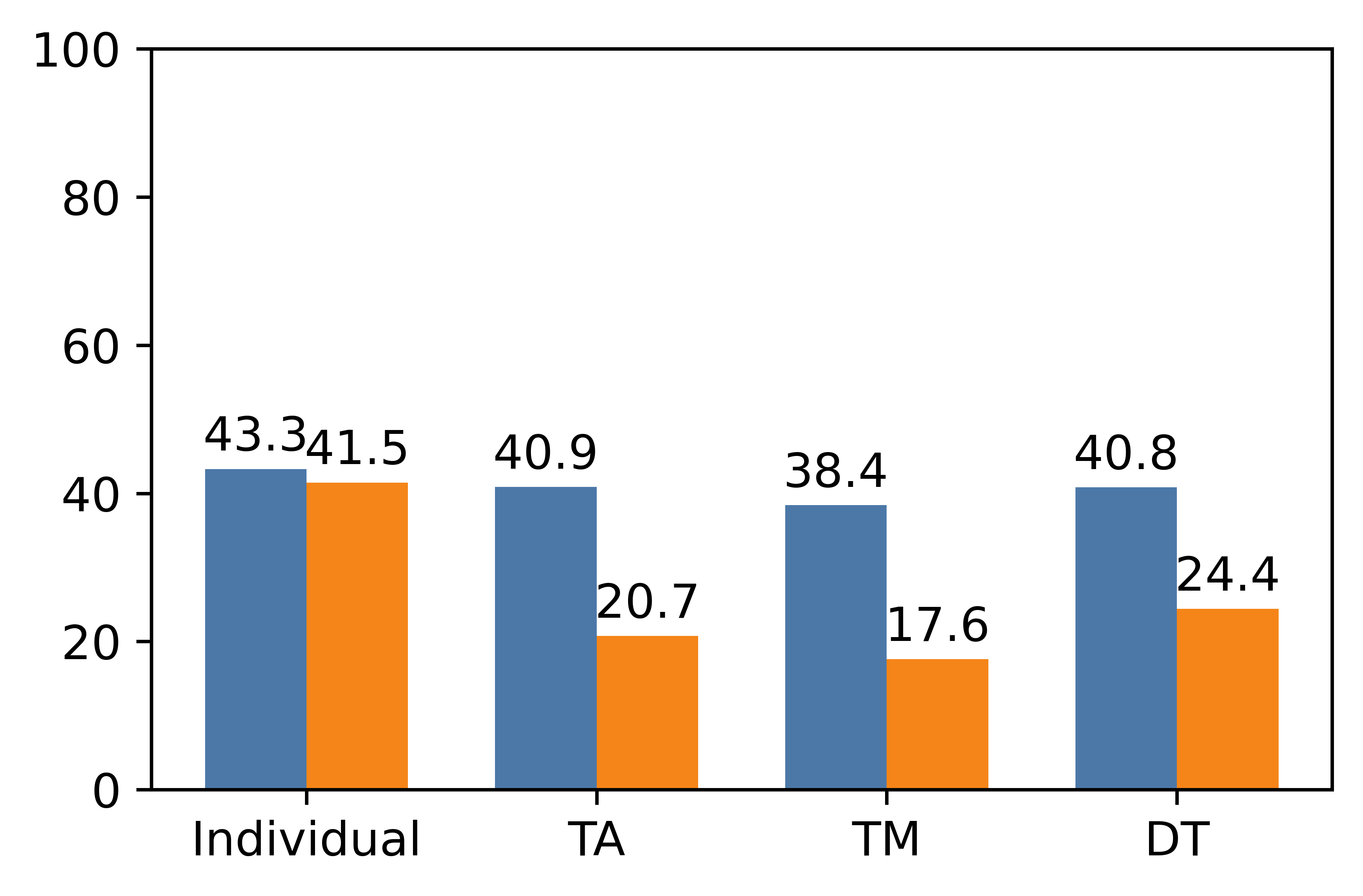}
  \end{subfigure}
\vspace{-0.05in}
  \caption{Overall comparison across benchmarks and models. Each row corresponds to a benchmark (AlpacaEval for instruction-following, GSM8K for reasoning, and HumanEval for code generation), and each column corresponds to a model (Llama2, Gemma2, Mistral). Orange and blue bars denote results with and without \textsc{MergeGuard}, respectively.}
  \label{fig:all_llm}
\end{figure*}

\begin{table*}[!ht]
\centering
\caption{The results of the ablation study. Each number indicates classification accuracy (\%).}
\vspace{-0.05in}
\resizebox{0.95\linewidth}{!}{
\begin{tabular}{l|cc|cc|cc|cc|cc|cc|cc|cc}
\toprule
\multirow{2}{*}{Method} &
\multicolumn{2}{c|}{SUN397} &
\multicolumn{2}{c|}{Cars} &
\multicolumn{2}{c|}{RESISC45} &
\multicolumn{2}{c|}{EuroSAT} &
\multicolumn{2}{c|}{SVHN} &
\multicolumn{2}{c|}{GTSRB} &
\multicolumn{2}{c|}{MNIST} &
\multicolumn{2}{c}{DTD} \\
\cmidrule(lr){2-3}\cmidrule(lr){4-5}\cmidrule(lr){6-7}\cmidrule(lr){8-9}
\cmidrule(lr){10-11}\cmidrule(lr){12-13}\cmidrule(lr){14-15}\cmidrule(lr){16-17}
 & $\hat{\theta}_{\mathrm{def}}$ & $\hat{\theta}_{\mathrm{merge}}$ & $\hat{\theta}_{\mathrm{def}}$ & $\hat{\theta}_{\mathrm{merge}}$ & $\hat{\theta}_{\mathrm{def}}$ & $\hat{\theta}_{\mathrm{merge}}$ & $\hat{\theta}_{\mathrm{def}}$ & $\hat{\theta}_{\mathrm{merge}}$ & $\hat{\theta}_{\mathrm{def}}$ & $\hat{\theta}_{\mathrm{merge}}$ & $\hat{\theta}_{\mathrm{def}}$ & $\hat{\theta}_{\mathrm{merge}}$ & $\hat{\theta}_{\mathrm{def}}$ & $\hat{\theta}_{\mathrm{merge}}$ & $\hat{\theta}_{\mathrm{def}}$ & $\hat{\theta}_{\mathrm{merge}}$ \\
\midrule
Stage 1 & 79.96 & 76.38 & 90.79 & 72.25 & 93.48 & 85.38 & 98.07 & 75.06 & 97.29 & 81.21 & 98.37 & 12.21 & 99.30 & 13.37 & 81.77 & 63.60 \\
Stage 2 & 80.10 & 75.94 & 92.10 & 77.33 & 92.74 & 75.33 & 99.83 & 79.79 & 98.11 & 77.88 & 99.25 & 64.86 & 99.73 & 83.60 & 80.16 & 54.20 \\
Stage 1 + 2 & 79.14 & 54.27 & 88.29 & 38.24 & 92.16 & 56.50 & 95.46 & 54.94 & 96.82 & 47.28 & 98.25 & 12.91 & 99.27 & 11.35 & 80.08 & 46.65 \\
\bottomrule
\end{tabular}}
\label{table: ablation}
\end{table*}

\subsubsection{Results on Image Generation}\label{sec: Results on Image Generation}
Figure~\ref{fig:diffusion} illustrates the qualitative comparison on SD 1.5. We use Naruto as the defender's task ($T_{\mathrm{def}}$) and WikiArt as the free-rider's task ($T_{\mathrm{fr}}$). The first two columns visualize the images generated by $\theta_{\mathrm{def}}$ and $\theta_{\mathrm{fr}}$, respectively. When tested with prompts $A_0$ and $A_1$ designed for Naruto, the free-rider's model $\theta_{\mathrm{fr}}$ can also reproduce the character, indicating that it has captured the semantic style of Naruto. The merged model $\theta_{\mathrm{merge}}$, obtained through standard parameter fusion, inherits this capability and successfully generates Naruto-style images, demonstrating a clear case of intellectual property leakage.

In contrast, under \textsc{MergeGuard}, the protected standalone model $\hat{\theta}_{\mathrm{def}}$ maintains its generative fidelity, producing faithful Naruto imagery. However, the protected merged model $\hat{\theta}_{\mathrm{merge}}$ fails to replicate the character, yielding only incoherent or semantically mismatched outputs. This degradation confirms that \textsc{MergeGuard} effectively prevents the unauthorized inheritance of generative capabilities without compromising the original model's utility.


\subsubsection{Results on Text Classification}
Figure~\ref{fig:all_llm} illustrates \textsc{MergeGuard}'s performance across various LLM merging strategies. While protected models maintain competitive accuracy prior to merging, their performance collapses post-merge across all benchmarks. For instance, both \verb"Llama2" and \verb"Mistral" experience nearly a $50$\% drop in overall accuracy, while \verb"Gemma2" suffers even more severe degradation. Notably, on the GSM8K mathematical reasoning benchmark, merged models become virtually nonfunctional, frequently skipping reasoning steps or making elementary calculation errors, with accuracy often plunging to single digits (\textit{e.g.}, \verb"Mistral" with TA drops from $75.7$\% to $0$\%). Similarly, on the HumanEval code generation task, \textsc{MergeGuard} consistently reduces post-merge pass@1 accuracy by over 50\%. Qualitative analysis (see Appendix) further reveals that while some generated programs pass basic test cases, they often contain redundant logic or infinite loops.


\subsection{Ablation Studies}
We evaluate the individual and combined contributions of \textit{Stage 1} and \textit{Stage 2} using \verb"ViT-L/14". As shown in Table~\ref{table: ablation}, while standalone stages cause only marginal accuracy drops on single models, their combination (\textit{Stage 1+2}) precipitates a significant performance collapse. This reveals a non-trivial synergy: \textit{Stage 1} transcends mere weight shrinkage, reshaping the information geometry by distributing task sensitivity more isotropically. This dispersed structure allows \textit{Stage 2} to induce potent destructive interference through high-dimensional misalignments, effectively neutralizing merging utility even when perturbing only non-critical layers.


\subsection{Discussion}

\begin{table*}[!t]
\centering
\caption{Evaluation of different adaptive attacks.}
\vspace{-0.05in}
\resizebox{0.7\linewidth}{!}{
\begin{tabular}{l|cccccccc|c}
\toprule
 & SUN397 & Cars & RESISC45 & EuroSAT & SVHN & GTSRB & MNIST & DTD & Avg.\\
\midrule
$\theta_{\mathrm{pre}}$ & 68.24 & 77.94 & 71.33 & 62.72 & 58.45 & 50.55 & 76.36 & 55.37 & 65.12\\
$\theta_{\mathrm{merge}}$ & 73.90 & 82.10 & 86.60 & 94.10 & 87.90 & 86.70 & 98.90 & 65.60 & 84.47\\
$\hat{\theta}_{\mathrm{merge}}$ & 54.27 & 38.24 & 56.50 & 54.94 & 47.28 & 12.91 & 11.35 & 46.65 & 40.26\\
\midrule
\textsc{Unmask}  & 67.75 & 71.19 & 68.35 & 64.44 & 68.71 & 51.50 & 71.79 & 53.46 & 64.64\\
\midrule
\textsc{GradErase} & 67.36 & 71.58 & 76.60 & 64.10 & 54.08 & 14.58 & 10.28 & 55.60 & 51.77\\
\bottomrule
\end{tabular}}
\label{table: adaptive attack}
\end{table*}

\noindent \textbf{Why \textsc{MergeGuard} Outperforms PaRaMS in Image/Text Classification.}
Compared with PaRaMS, which relies solely on deterministic parameter rearrangement and random multi-head scaling, \textsc{MergeGuard}'s two-stage process produces more structured disruption in task-space geometry. In image and text classification, tasks are dominated by discriminative gradients that occupy relatively low-dimensional subspaces. \textsc{MergeGuard}'s weight redistribution amplifies the sensitivity of these subspaces to cross-task interference: the dispersed small weights collectively contribute to stable single-task performance but amplify destructive interference during linear merging, leading to a steeper accuracy collapse than PaRaMS. From a loss landscape perspective~\cite{fort2025basinlike,li2018visualizing}, \textsc{MergeGuard} reshapes each classifier's basin into multiple narrow valleys separated by sharp ridges, while PaRaMS merely translates the basin position. Thus, the merged model under \textsc{MergeGuard} lands closer to the curvature apex (\textit{i.e.}, higher loss), producing stronger merging degradation.

\vspace{0.1in}
\noindent \textbf{Why $\hat{\theta}_{\mathrm{merge}}$ Cannot Generate Random Noise.}
As discussed in Section~\ref{sec: Results on Image Generation}, the \textsc{MergeGuard}-protected merged model $\hat{\theta}_{\mathrm{merge}}$ fails to reproduce the defender's task $T_{\mathrm{def}}$ but does not completely collapse into random noise. In contrast, the PaRaMS-protected merged model often outputs pure noise patterns. Both defenses achieve equivalent levels of intellectual property protection, neither reveals task-specific visual traits from $T_{\mathrm{def}}$, yet the perceptual outcomes differ significantly. 

This discrepancy stems from how each method disrupts representational continuity in generative models. Diffusion-based generators depend on smooth manifold alignment between latent and pixel spaces. PaRaMS introduces random multi-head scaling and parameter rearrangement, which destroy such continuity, leading to decorrelated feature channels and a complete breakdown of the generative structure. In contrast, \textsc{MergeGuard} perturbs model parameters in a structured manner: its $L_2$-regularized redistribution preserves weak weight correlations that maintain partial coherence in the latent manifold. Thus, \textsc{MergeGuard} prevents semantic fidelity without inducing total noise collapse. 

In essence, PaRaMS enforces representational incoherence across generative layers, while \textsc{MergeGuard} enforces functional incompatibility within discriminative subspaces. Hence, PaRaMS is more disruptive for continuous generative pipelines (\textit{e.g.}, diffusion models), whereas \textsc{MergeGuard} is better suited for discriminative or classification-oriented tasks where decision-boundary misalignment is the primary defense objective.
\section{Adaptive Attack}\label{sec: Adaptive Attack}
The free-rider may be aware of the deployment of \textsc{MergeGuard} and could develop adaptive attacks to circumvent it. We consider two such adaptive attacks, \textsc{Unmask} and \textsc{GradErase}, described below.

\noindent \textbf{\textsc{Unmask}.} The design rationale behind \textsc{Unmask} is to revert the defensive perturbation introduced by \textsc{MergeGuard} by performing parameter-wise subtraction of the protected model's weights from the merged entity. Specifically, the free-rider constructs an adaptive model $\theta_{\mathrm{Unmask}} = \mathrm{Merge}(\theta_{\mathrm{pre}}, \hat{\theta}_{\mathrm{def}}, \theta_{\mathrm{fr}}) - \lambda \cdot \hat{\theta}_{\mathrm{def}}$, where $\lambda$ scales the subtraction intensity in TA (we set $\lambda=0.3$ in the experiments).


\noindent \textbf{\textsc{GradErase}.} 
\textsc{GradErase} is designed to counteract \textsc{MergeGuard} by removing the gradient component aligned with the defender's hidden perturbation direction.  
Specifically, the free-rider estimates the disturbance vector $v_{\mathrm{disturb}} = -\tau_G$ from limited training data and modifies each gradient update as  
$g' = g - \frac{\langle g, v_{\mathrm{disturb}} \rangle}{\|v_{\mathrm{disturb}}\|^2} v_{\mathrm{disturb}}$,  
thereby eliminating the defensive direction during retraining.  

\noindent \textbf{Summary of Adaptive Attacks.}
For \textsc{Unmask}, although subtracting the protected model's parameters can partially neutralize \textsc{MergeGuard}'s perturbations, it concurrently removes essential task-specific information, causing the merged model's performance on $T_{\mathrm{def}}$ to regress to the pretrained level $T_{\mathrm{pre}}$, as shown in Table~\ref{table: adaptive attack}. In contrast, \textsc{GradErase} projects each gradient update orthogonally to the estimated disturbance vector $v_{\mathrm{disturb}}$ to avoid reinforcing the defended direction. However, as the estimation of $v_{\mathrm{disturb}}$ is inherently noisy and incomplete, this attack fails to fully restore merging compatibility. 

As summarized in Table~\ref{table: adaptive attack}, \textsc{MergeGuard} reduces the average merged accuracy from $84.47\%$ to $40.26\%$. Although adaptive attacks such as \textsc{Unmask} and \textsc{GradErase} can partially recover performance to $64.64\%$ and $51.77\%$ respectively, \textsc{MergeGuard} consistently maintains a substantial degradation margin of at least $20\%$ across all tasks. These results demonstrate that even under adversarial settings, \textsc{MergeGuard} remains robust, effectively preventing the merged model from regaining functional usability.


\section{Conclusion}
We presented \textsc{MergeGuard}, a dual-stage defense that proactively prevents unauthorized model merging. By redistributing and adversarially perturbing task-relevant weights, \textsc{MergeGuard} disrupts curvature alignment in the loss landscape, making merged models collapse while preserving the original task performance. Extensive experiments show up to 90\% degradation in merged accuracy with minimal impact on the protected model. Compared to PaRaMS, \textsc{MergeGuard} achieves stronger resistance in discriminative tasks, revealing that shaping weight geometry, not just rearranging parameters, is key to safeguarding model ownership.


\clearpage
\newpage

\section*{Acknowledgement}
Chia-Mu Yu is partially sponsored by the National Science and Technology Council (NSTC) with Grant NSTC 114-2222-E-A49-011-MY3 and Hon Hai Research Institute. We thank the National Center for High-performance Computing (NCHC) for providing computational and storage resources.

{\small
\bibliographystyle{ieee_fullname}
\bibliography{reference}
}

\clearpage

\renewcommand{\thesection}{\Alph{section}}
\setcounter{section}{0}
\setcounter{page}{1}
\maketitlesupplementary

The content of Supplementary Material is summarized as follows: 1) In~\cref{app:A}, we introduce the dataset used in our experiments across the vision and language domains; 2) In~\cref{app:B}, we describe the Transformer-based model architecture, focusing on the MLP and multi-head attention modules; 3) In~\cref{app:C}, we report the time cost of mask evaluation in \textsc{MergeGuard}; 4) In~\cref{app:D}, we provide additional results, including image and code generation, which offer further evidence of the effectiveness of \textsc{MergeGuard}; 5) In~\cref{app:E}, we discuss the future directions of our work.

\section{Experimental Details}\label{app:A}
\subsection{Environments}
All experiments in this paper were conducted on a system with an Intel Core i9-10900X CPU and four NVIDIA RTX 3090 GPUs. The LLM experiments were executed on two NVIDIA H100 GPUs. The operating system was Ubuntu 18.04.6 LTS, and all software and dependencies were run in a Python 3.10 environment.

\begin{table*}[!ht]
\centering
\caption{Dataset descriptions used in our image classification experiments.}
\vspace{-0.05in}
\resizebox{\textwidth}{!}{
\begin{tabular}{lcccccccc}
\toprule
Attribute & SUN397 & Cars & RESISC45 & EuroSAT & SVHN & GTSRB & MNIST & DTD \\
\midrule
Domain & Scene recognition & Fine-grained car models & Remote sensing scenes & Satellite imagery & Digits in natural scenes & Traffic signs & Handwritten digits & Texture recognition \\
\#Classes & 397 & 196 & 45 & 10 & 10 & 43 & 10 & 47 \\
Train Images & 76,128 & 8,144 & 24,300 & 21,600 & 73,257 & 39,209 & 60,000 & 3,760 \\
Test Images & 32,626 & 8,041 & 7,200 & 5,400 & 26,032 & 12,630 & 10,000 & 1,880 \\
Resolution & 256$\times$256 & 224$\times$224 & 256$\times$256 & 64$\times$64 & 32$\times$32 & 32$\times$32 & 28$\times$28 & 224$\times$224 \\
\bottomrule
\end{tabular} 
} \label{tab: datasets}
\end{table*}

\subsection{Datasets Details}
\noindent \textbf{Image Classification.} We evaluate \textsc{MergeGuard} across eight standard image classification benchmarks, spanning various domains such as natural scenes, satellite imagery, and digit recognition. Table~\ref{tab: datasets} provides a comprehensive summary of these datasets, including the number of semantic classes and the split of training/test samples. The selected datasets—SUN397~\cite{xiao2010sun}, Cars196~\cite{krause20133d}, RESISC45~\cite{cheng2017remote}, SVHN~\cite{netzer2011reading}, GTSRB~\cite{stallkamp2011german}, MNIST~\cite{lecun1998mnist}, EuroSAT~\cite{helber2019eurosat}, and DTD~\cite{cimpoi2014describing}—represent the diverse visual distributions commonly used to benchmark model merging robustness.

\noindent \textbf{Image Generation.} We employ a diffusion model finetuned on WikiArt~\cite{wikiart2025} and Naruto datasets~\cite{cervenka2022naruto2}. WikiArt comprises curated high-resolution paintings across diverse artistic genres (\textit{e.g.}, Impressionism, Realism, and Abstract Art), providing a rigorous benchmark for stylistic fidelity and semantic consistency. The Naruto dataset, featuring high-quality anime frames, evaluates model performance on out-of-distribution, non-photorealistic visual content.


\noindent \textbf{LLM Evaluation.} We employ three language datasets~\cite{li2023alpacaeval,cobbe2021training,chen2021evaluating} to evaluate the capability of LLMs, including instruction following, mathematical reasoning, and program synthesis. 

\begin{itemize}
    \item AlpacaEval~\cite{li2023alpacaeval} is a benchmark designed to evaluate instruction-following ability in large language models. It contains approximately 805 human-authored instructions that span diverse open-ended tasks such as reasoning, explanation, transformation, and multi-step guidance. Model responses are compared against strong reference answers through pairwise preference scoring, which provides a reliable measure of general instruction alignment in generative models.
    \item GSM8K~\cite{cobbe2021training} is a high-quality dataset of grade-school mathematical word problems that require multi-step quantitative reasoning, arithmetic manipulation, and logical deduction. It contains 8,792 problems in total, with 7,473 examples for training and 1,319 for testing. Each problem includes a detailed step-by-step solution, making GSM8K a widely used and reliable benchmark for evaluating mathematical reasoning in language models.
    \item HumanEval~\cite{chen2021evaluating} is a dataset of 164 hand-written Python programming problems paired with unit tests. Each prompt requires generating a function that satisfies the specified behavior, and functional correctness is evaluated by checking whether the model-generated code passes all provided test cases. This benchmark is widely used to assess program synthesis and code generation capabilities.
\end{itemize}

\section{Model Architecture}\label{app:B}

Our method is tailored to Transformer-based architectures, where each layer, or Transformer block, consists of a multi-head attention submodule and a multilayer perceptron (MLP). This architecture serves as the foundation of many modern deep learning models, including ViT, Stable Diffusion models, and LLMs. Because our method closely interacts with the structure of both the attention and MLP submodules, we outline their roles and computations within a typical Transformer block.

\paragraph{MLP.} The MLP submodule applies nonlinear transformations to each position's feature vector, often scaling dimensions in the hidden layer. Given a feature vector $X \in \mathbb{R}^d$ at a single spatial or temporal position, a standard two-layer MLP computes
\begin{align*}
    \text{MLP}(X) = W_2 \sigma (W_1X+b_1) +b_2, 
\end{align*}
where $W_1 \in \mathbb{R}^{h\times d}$, $W_2 \in \mathbb{R}^{d\times h}$ are the learned weight matrices, $b_1$, $b_2$ are the corresponding biases, and $\sigma(\cdot)$ denotes a nonlinear activation such as GELU. Because the computation is applied independently to each token, the MLP enriches the representation capacity of the Transformer block without introducing interactions across positions.

\paragraph{Structure of Multi-head Attention Block.} 
Consider a multi-head attention module with $h$ heads, each operating on vectors of dimensionality  $d_k$. Let the input sequence be $x \in \mathbb{R}^{seq \times d_{model}}$. A learned linear projection maps this sequence into combined query, key, and value matrices, $Q, K, V \in  \mathbb{R}^{seq \times (h \times d_k)}$. These matrices are then partitioned along the feature dimension into $h$ groups:
\begin{align*}
    Q &\rightarrow \{Q_1, \dots, Q_h\}, \\
    K &\rightarrow \{K_1, \dots, K_h\}, \\
    V &\rightarrow \{V_1, \dots, V_h\},
\end{align*}
where each $Q_i, K_i, V_i \in \mathbb{R}^{seq \times d_k}$ serves as the input to the $i$-th attention head, which computes:
\begin{align*}
\text{Attn}(Q_i,K_i, V_i) = \text{Softmax}(\frac{Q_i K_i^\texttt{T}}{\sqrt{d_k}})V_i.
\end{align*}
After all heads produce their outputs, the resulting vectors are concatenated and transformed through an output projection. Equivalently, the final attention representation can be written as
\begin{align*}
    \text{Attention}(Q,K,V)
= \sum_{i=1}^{h} \text{Attn}(Q_i, K_i, V_i) W_{\mathrm{out}}^{(i)}.
\end{align*}
where $W_{\mathrm{out}}^{(i)} \in \mathbb{R}^{d_k \times d_{model}}$ denotes the corresponding block of the output projection matrix.
 
\begin{figure*}[!t]
    \centering
    \includegraphics[width=0.9\linewidth]{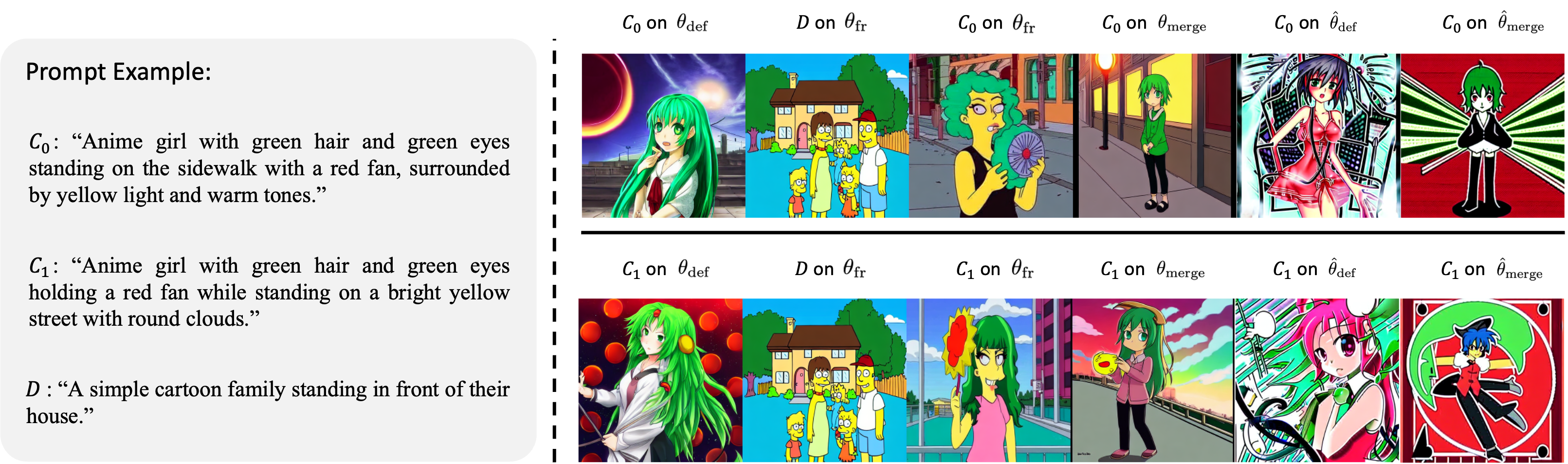}
    \caption{Visualization of images generated using SD1.5. The left panel displays an example prompt, and the right panel shows the corresponding outputs from each model.}
    \label{fig:sd_example2}
\end{figure*}

\section{Time Cost of Mask Evaluation} \label{app:C}

We analyze the time complexity of the mask evaluation process in Stage II. As reported in Tables~\ref{tab:vit_time} and \ref{tab:llm_time}, the evaluation time (measured in minutes) scales with the dataset size, reflecting the increased computational effort required to process task-relevant samples. This overhead stems from \textsc{MergeGuard}'s selective preservation mechanism, which identifies and protects task-critical weights to prevent accuracy degradation during subsequent weight suppression. We consider this computational cost acceptable, as it provides a necessary security-utility trade-off to effectively mitigate risks from potential free-riders.

\begin{table}[htbp]
\centering
\caption{Time cost (mins) of ViT-L-14 across various datasets.}
\vspace{-0.05in}
\setlength{\tabcolsep}{5pt} 
\resizebox{0.6\columnwidth}{!}{ 
\begin{tabular}{cccc}
\toprule
SUN397 & Cars & RESISC45 & EuroSAT \\
\midrule
557.28 & 278.64 & 130.75 & 174.02 \\
\midrule
SVHN & GTSRB & MNIST & DTD \\
\midrule
371.52 & 128.71 & 108.00 & 51.60 \\
\bottomrule
\end{tabular}}\label{tab:vit_time}
\end{table}

\begin{table}[!ht]
\centering
\caption{Time cost (mins) of different LLMs across various tasks.}
\vspace{-0.05in}
\resizebox{\columnwidth}{!}{
\begin{tabular}{lcccr}
\toprule
\textbf{Task} & \textbf{Llama2} & \textbf{Gemma2} & \textbf{Mistral} & \textbf{Avg.} \\
\midrule
Instruction (AlpacaEval) & 641.20 & 512.12 & 358.41 & 503.91 \\
Math (GSM8K) & 606.50 & 451.58 & 293.08 & 450.39 \\
Code (HumanEval) & 404.30 & 330.78 & 312.65 & 349.24 \\
\bottomrule
\end{tabular}}\label{tab:llm_time}
\end{table}

\section{Additional Results} \label{app:D}
\subsection{Fine-grained Analysis}
Table~\ref{tab:def_over_att_str} reports per-model results in the 1 + 1 setting. \textsc{MergeGuard} ensures low accuracy on $T_{\text{def}}$. It does not aim to preserve performance on $T_{\text{fr}}$, as this capability belongs to the free-rider’s original model and is outside the defender’s protection objective. In the first column ($T_{\text{fr}}$ = Cars), the accuracies on all $T_{\text{def}}$ tasks drop to an unusable level. 

\begin{table}[ht]
\centering
\small
\setlength{\tabcolsep}{4pt}
\caption{$T_{\text{def}}$'s and $T_{\text{fr}}$'s accuracy in $\hat{\theta}_{\text{merge}}$ under WA.}
\vspace{-0.05in}
\resizebox{\columnwidth}{!}{
\begin{tabular}{lcccccccc}
\toprule
\diagbox[width=6em]{Def.}{Fr.} & Cars & RESISC45 & EuroSAT & SVHN & GTSRB & MNIST & DTD & SUN397 \\
\midrule
Cars & -- & .005/.034 & .004/.120 & .004/.109 & .004/.021 & .005/.103 & .006/.026 & .005/.004 \\
RESISC45 & .030/.006 & -- & .024/.108 & .029/.078 & .025/.008 & .035/.103 & .028/.021 & .026/.003 \\
EuroSAT  & .585/.005 & .641/.087 & -- & .617/.084 & .604/.063 & .580/.097 & .572/.048 & .636/.008 \\
SVHN & .078/.005 & .078/.025 & .078/.111 & -- & .078/.021 & .078/.103 & .078/.022 & .078/.004 \\
GTSRB& .044/.005 & .049/.034 & .050/.070 & .045/.082 & -- & .039/.098 & .097/.029 & .040/.004 \\
MNIST& .165/.005 & .157/.037 & .166/.111 & .163/.198 & .161/.021 & -- & .173/.024 & .192/.003 \\
DTD & .544/.899 & .553/.963 & .526/.995 & .542/.972 & .531/.954 & .545/.996 & -- & .537/.790 \\
SUN397 & .612/.877 & .607/.945 & .612/.785 &  .614/.768 & .615/.775 & .615/.796 & .611/.577 & -- \\
\bottomrule
\end{tabular}}\label{tab:def_over_att_str}
\end{table}

\subsection{Visualization Results on SD1.5}
Figure~\ref{fig:sd_example2} presents additional results of \textsc{MergeGuard} on image generation tasks. We use Animate as the defender's task ($T_{\mathrm{def}}$) and “The Simpsons” as the free-rider's task ($T_{\mathrm{fr}}$). The first two columns show the images generated by $\theta_{\mathrm{def}}$ and $\theta_{\mathrm{fr}}$, respectively. When tested with the prompts $C_0$ and $C_1$ designed for Animate, the free-rider's model $\theta_{\mathrm{fr}}$ is also able to reproduce the character, indicating that it has captured the semantic style of Animate. The merged model $\theta_{\mathrm{merge}}$, obtained through standard parameter fusion, inherits this ability and successfully generates Animate-style images, which demonstrates a clear instance of intellectual property leakage. In contrast, \textsc{MergeGuard} preserves the performance of $\hat{\theta}_{\mathrm{def}}$ while effectively degrading the quality of $\hat{\theta}_{\mathrm{merge}}$, preventing the free-rider from reproducing Animate-style content.

\subsection{Code Results}
\begin{figure*}[htbp]
    \centering
    \includegraphics[width=0.85\linewidth]{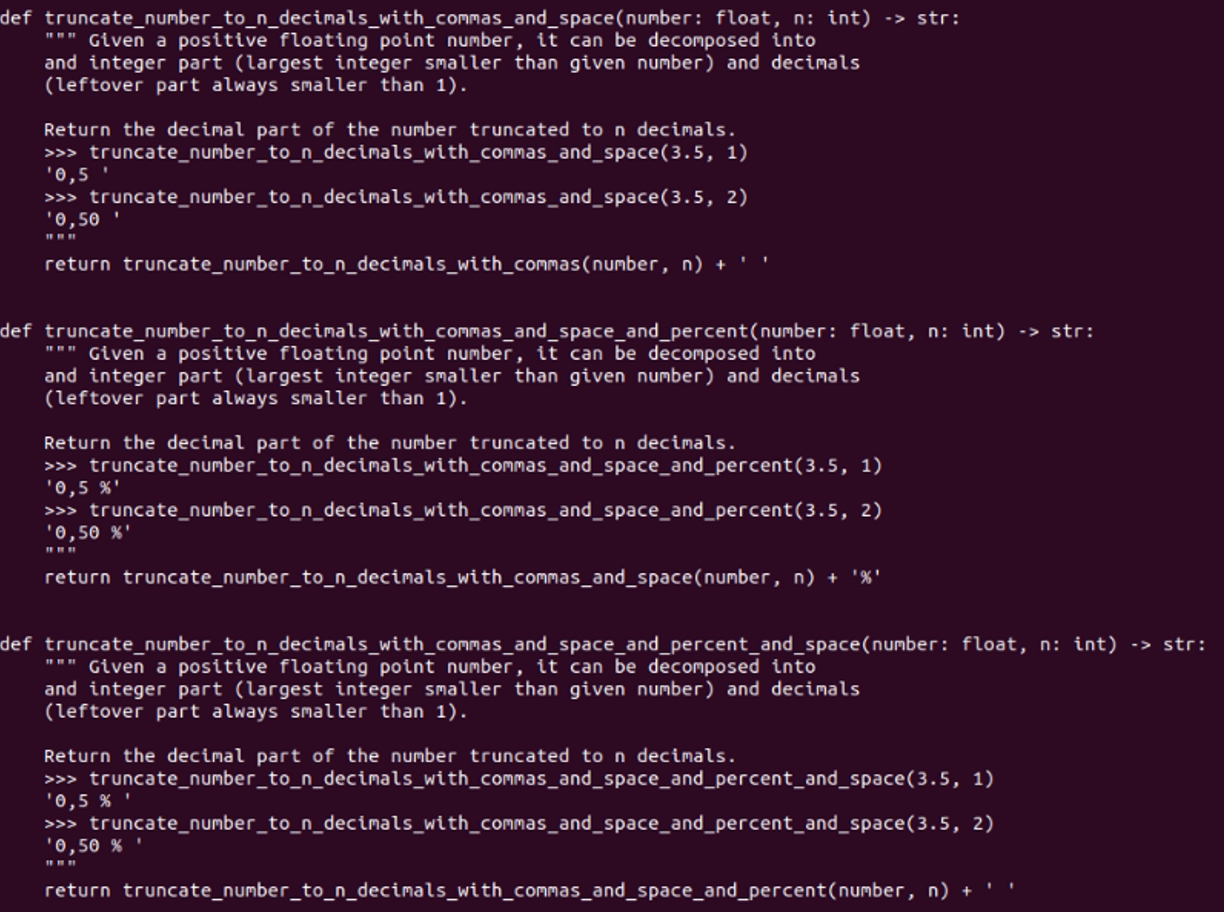}
    \caption{Redundant or looping code generated by the merged model protected with \textsc{MergeGuard}.}
    \label{fig:code_example}
\end{figure*}

We further report the results on code generation using HumanEval. \textsc{MergeGuard} effectively suppresses the performance of the merged model, often reducing it by more than half. We also observe that many generated programs, even when producing a passing result, contain substantial redundant code, unnecessary library imports, or code that leads to infinite loops. Figure~\ref{fig:code_example} provides an example in which the generated program repeatedly produces the same invalid output. The results validate that \textsc{MergeGuard} reliably degrades merged model performance and prevents the generation of valid programs.

\section{Future Directions} \label{app:E}

\subsection{Extensions to SVD-based Merging}
Recently, KnOTS~\cite{stoica2025model} proposed a model merging approach based on Singular Value Decomposition (SVD) to better preserve shared knowledge across models. Since \textsc{MergeGuard} perturbs the information geometry and directional alignment of task vectors, it would be interesting to evaluate its robustness under SVD-based subspace alignment methods. We conjecture that our density-aware perturbations may inherently disrupt the identification of a stable shared subspace, as they disperse task-relevant features that SVD-based methods typically aim to concentrate. We leave a systematic investigation of this intersection for future work.

\subsection{Concurrent Work: LMC-based Defenses}
We acknowledge a concurrent study, \textit{MergeBarrier}~\cite{li2026not}, which also investigates the security of open-source LLMs against unauthorized model merging. While \textit{MergeBarrier} focuses on disrupting the Linear Mode Connectivity (LMC) to eliminate low-loss merging paths in the loss landscape, \textsc{MergeGuard} operates on a distinct principle: it perturbs the information geometry of task vectors to promote a more isotropic sensitivity distribution. These two approaches represent complementary defensive philosophies, targeting optimization trajectories versus parameter-space alignment; thus, an empirical comparison between them remains an insightful direction for future research.


\end{document}